\documentclass{article}

\usepackage[preprint]{neurips_2025}

\usepackage[utf8]{inputenc} 
\usepackage[T1]{fontenc}    
\usepackage{hyperref}       
\usepackage{url}            
\usepackage{booktabs}       
\usepackage{amsfonts}       
\usepackage{amsmath}        
\usepackage{nicefrac}       
\usepackage{microtype}      
\usepackage{xcolor}         
\usepackage{algorithm}
\usepackage{algorithmicx,algpseudocode}
\usepackage{wrapfig}
\usepackage{graphicx}
\usepackage{subfigure}
\usepackage{makecell}

\title{Exploring Polyglot Harmony: On Multilingual Data Allocation for  Large Language Models Pretraining}

\author{%
  Ping Guo\textsuperscript{\ddag}\quad\quad
  Yubing Ren\textsuperscript{\dag}\quad\quad
  Binbin Liu\textsuperscript{\ddag} \quad\quad
  Fengze Liu\textsuperscript{\ddag} \\
  \textbf{Haobin Lin\textsuperscript{\ddag}\quad
  Yifan Zhang\textsuperscript{\ddag}\quad
  Bingni Zhang\textsuperscript{\ddag}\quad
  Taifeng Wang\textsuperscript{\ddag}\quad
  Yin Zheng\textsuperscript{\ddag}}\\
\textsuperscript{\ddag}ByteDance \quad\quad
\textsuperscript{\dag}Institute of Information Engineering, Chinese Academy of Sciences
}

\begin{document}

\maketitle

\begin{abstract}
Large language models (LLMs) have become integral to a wide range of applications worldwide, driving an unprecedented global demand for effective multilingual capabilities. Central to achieving robust multilingual performance is the strategic allocation of language proportions within training corpora. However, determining optimal language ratios is highly challenging due to intricate cross-lingual interactions and sensitivity to dataset scale. This paper introduces \textsc{Climb} (\textbf{C}ross-\textbf{L}ingual \textbf{I}nteraction-aware \textbf{M}ultilingual \textbf{B}alancing), a novel framework designed to systematically optimize multilingual data allocation. At its core, \textsc{Climb} introduces a \textit{cross-lingual interaction-aware language ratio}, explicitly quantifying each language’s effective allocation by capturing inter-language dependencies. Leveraging this ratio, \textsc{Climb} proposes a principled two-step optimization procedure—first equalizing marginal benefits across languages, then maximizing the magnitude of the resulting language allocation vectors—significantly simplifying the inherently complex multilingual optimization problem. Extensive experiments confirm that \textsc{Climb} can accurately measure cross-lingual interactions across various multilingual settings. LLMs trained with \textsc{Climb}-derived proportions consistently achieve state-of-the-art multilingual performance, even achieve competitive performance with open-sourced LLMs trained with more tokens.
\end{abstract}

\section{Introduction}

Large language models (LLMs), exemplified by the GPT series~\citep{openai2024gpt4technicalreport,openai2024gpt4ocard}, LLaMA series~\citep{touvron2023llamaopenefficientfoundation,touvron2023llama2openfoundation,grattafiori2024llama3herdmodels}, Gemma series~\citep{gemmateam2024gemmaopenmodelsbased,gemmateam2024gemma2improvingopen,gemmateam2025gemma3technicalreport}, Qwen series~\citep{bai2023qwentechnicalreport,qwen2025qwen25technicalreport,yang2025qwen3technicalreport}, and DeepSeek series~\citep{deepseekai2025deepseekv3technicalreport,deepseekai2025deepseekr1incentivizingreasoningcapability}, have reshaped various language-based applications worldwide, powering advanced chatbots \cite{dam2024completesurveyllmbasedai}, machine translation systems \cite{zhu-etal-2024-multilingual}, and intelligent virtual assistants \cite{10890139}. Such impressive capabilities emerge predominantly from extensive pretraining on enormous textual datasets, frequently spanning tens to hundreds of trillions of tokens, enabling the capture of rich and diverse linguistic knowledge. Driven by the growing global demand and the need for equitable language representation, there has been an accelerating shift toward multilingual pretraining, aiming to transcend linguistic boundaries and serve a broader range of linguistic communities effectively \cite{zhu2024multilinguallargelanguagemodels}. Central to this shift lies a fundamental question: \textbf{how should the proportions of different languages be optimally allocated within the training corpus to achieve balanced and superior model performance across all target languages?}

\begin{wrapfigure}{r}{0.4\textwidth}
    \centering
    \includegraphics[width=0.38\textwidth]{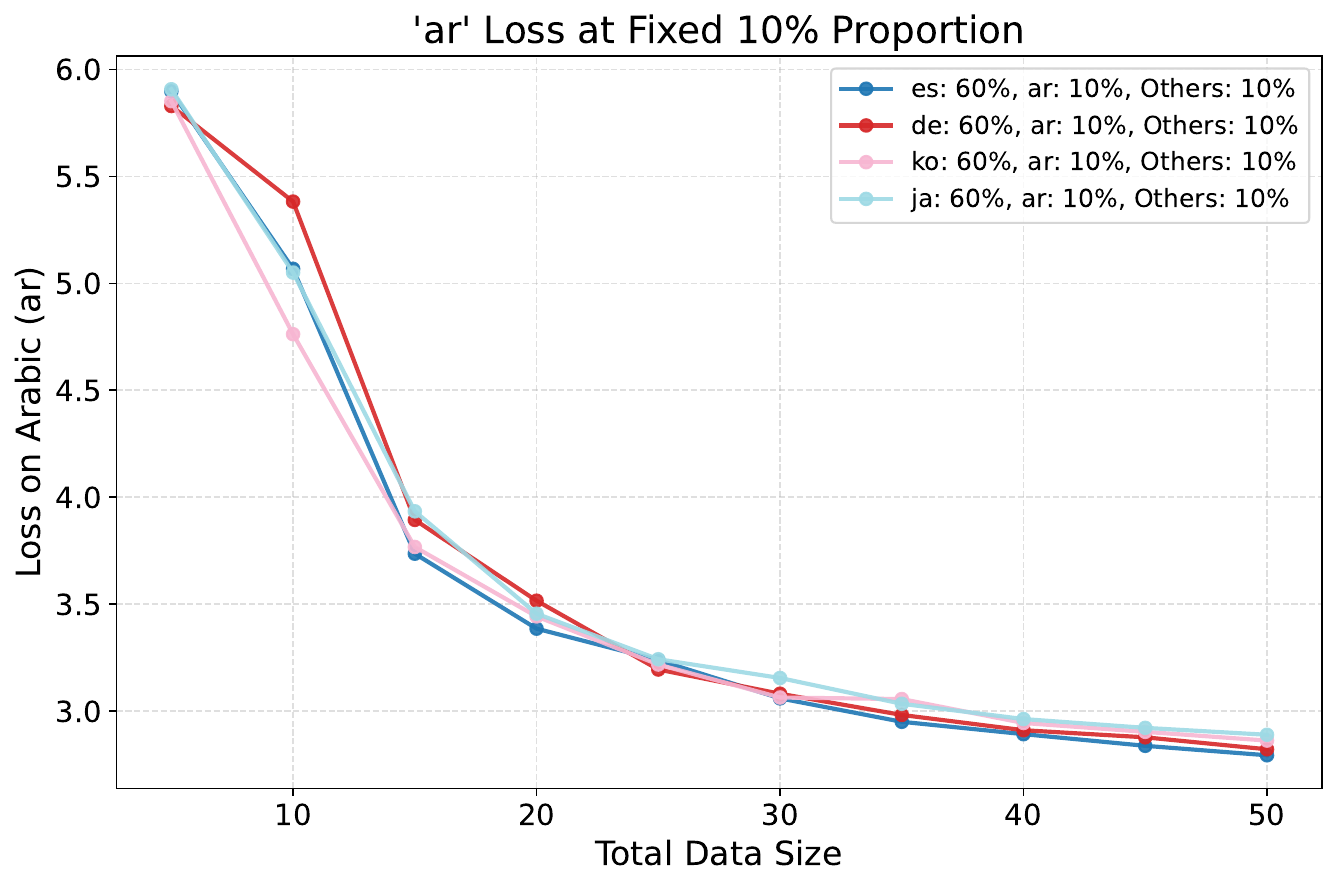}
    \caption{Cross-lingual Interactions in a five-langauge LLM.}
    \label{fig:ci_prove}
\end{wrapfigure}

However, determining an optimal multilingual mixture poses considerable challenges. The foremost difficulty arises from cross-lingual interactions: performance on one language can be significantly influenced by other languages trained concurrently \cite{faisal-anastasopoulos-2024-efficient,choenni-etal-2023-languages}. As illustrated in Figure~\ref{fig:ci_prove}, even when the training proportion of Arabic remains fixed to 10\%, modifying the proportions of the other four languages (increasing one language to 60\% proportion) in a five-language LLM can substantially alter Arabic's performance. This interdependence prevents isolated optimization of individual languages and necessitates joint optimization of the entire language set. Additionally, optimal language ratios are sensitive to the scale of the training corpus \cite{kang2025autoscalescaleawaredatamixing,he2024scalinglawsmultilinguallanguage,10.5555/3600270.3601689,goyal2024the}. Specifically, language proportions identified as optimal at smaller scales (e.g., 1 billion tokens) may no longer remain optimal when scaled to larger training sets (e.g., 4 trillion tokens), rendering simple extrapolations unreliable and incurring prohibitive experimental costs. Consequently, current multilingual LLMs often resort to heuristic trial-and-error approaches~\citep{rae2022scalinglanguagemodelsmethods,grattafiori2024llama3herdmodels}, or reuse language ratios derived from prior models without systematic justification~\citep{lai-etal-2024-llms}, highlighting a critical need for a principled and scalable solution to multilingual data allocation.

In pursuit of achieving the optimal language allocation, this paper explores whether it is possible to accurately predict model performance under various language allocations without explicitly training the models. Inspired by the concept of scaling laws, which characterize how a model's validation loss systematically scales with model size ($N$) and data volume ($D$)~\citep{kaplan2020scalinglawsneurallanguage,10.5555/3600270.3602446}, we hypothesize that a similar predictive framework could be applied to multilingual settings by incorporating language proportions. Specifically, if we can formulate a mathematical relationship that captures how validation performance varies with language proportions in the training corpus, then it becomes feasible to infer optimal language ratios by identifying the allocations that minimize validation loss. However, due to the intricate cross-lingual interactions among languages, precisely modeling and predicting validation performance across different language compositions remains highly challenging.

In this paper, we propose \textbf{\textsc{Climb} (Cross-Lingual Interaction-aware Multilingual Balancing)}, a novel framework designed to systematically optimize language proportions for multilingual LLM pre-training. Our approach consists of two interconnected components. First, we introduce the \textit{cross-lingual interaction-aware language ratio}, a novel metric that explicitly quantifies the effective allocation of each language in the presence of cross-lingual interactions, effectively reflecting the impact of other jointly trained languages. Second, leveraging these cross-lingual interaction-aware ratios, we can estimate the optimal multilingual balance by decomposing the optimization into two steps: initially, we determine the direction of optimal allocation by equalizing the marginal benefits across languages; subsequently, we obtain the estimated optimal proportions by maximizing the magnitude of the resulting cross-lingual interaction-aware language ratio vector. This principled two-step procedure enables efficient and accurate computation of multilingual data distributions, significantly reducing the complexity inherent in direct joint optimization.

To comprehensively evaluate the effectiveness of \textsc{Climb}, we conduct experiments in two primary aspects. First, we validate the predictive accuracy of the proposed cross-lingual interaction-aware language ratio. By integrating this novel ratio into the multilingual scaling law framework, we observe a substantial improvement in predictive accuracy compared to baseline scaling laws relying on independence assumptions among languages. Second, leveraging the optimal proportions computed via \textsc{Climb}, we train multilingual LLMs at both 1.2B and 7B parameter scales. Experimental results demonstrate that models pretrained with \textsc{Climb}-derived ratios consistently achieve state-of-the-art performance compared to various baselines with alternative language allocations. Remarkably, even compared to open-sourced models pretrained on more tokens, our \textsc{Climb}-optimized models exhibit highly competitive performance across multiple multilingual benchmarks.

    
    

\section{\textsc{Climb}}
Our approach comprises two main components: the \emph{Cross-lingual Interaction-aware Language Ratio}, which explicitly quantifies effective language proportions by considering cross-lingual interactions, and the \emph{Optimal Multilingual Balance}, which leverages these interaction-aware ratios to estimate the optimal language allocation $\mathbf{r}^*$ that minimizes multilingual validation loss.



\subsection{Problem Formulation}

Given a multilingual corpus consisting of training data from $m$ distinct languages ${L_1, \dots, L_m}$, our objective is to determine the optimal language allocation for pretraining LLMs. Formally, we define the language proportion vector as $\mathbf{r} = [r_1, r_2, \dots, r_m]^\top \in \mathcal{R}^m$, where $\mathcal{R}^m = \{ \mathbf{r} \in \mathbb{R}^m \mid \sum_{i=1}^{m} r_i = 1, r_i \geq 0, \forall i \}$ denotes the probability simplex.

\begin{figure}[t]
    \centering
    \subfigure[$\tilde{r}_{ar}$ as a function of $r_{ar}$ with varying co-training languages]{
        \includegraphics[width=0.3\textwidth]{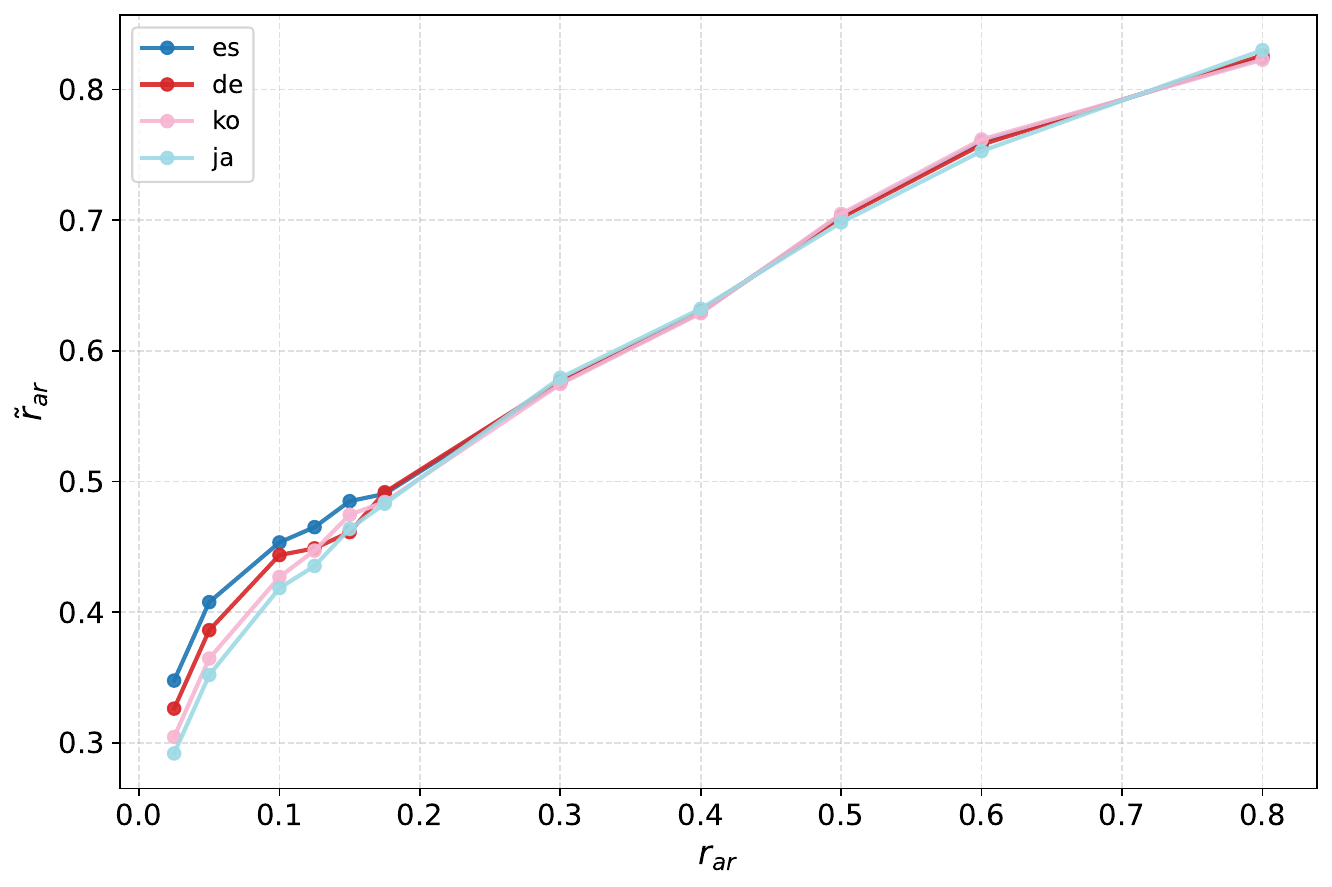}
    }\hfill
    \subfigure[$\tilde{r}_{ar}$ as a function of $r_{ar}$ with varying language counts]{
        \includegraphics[width=0.3\textwidth]{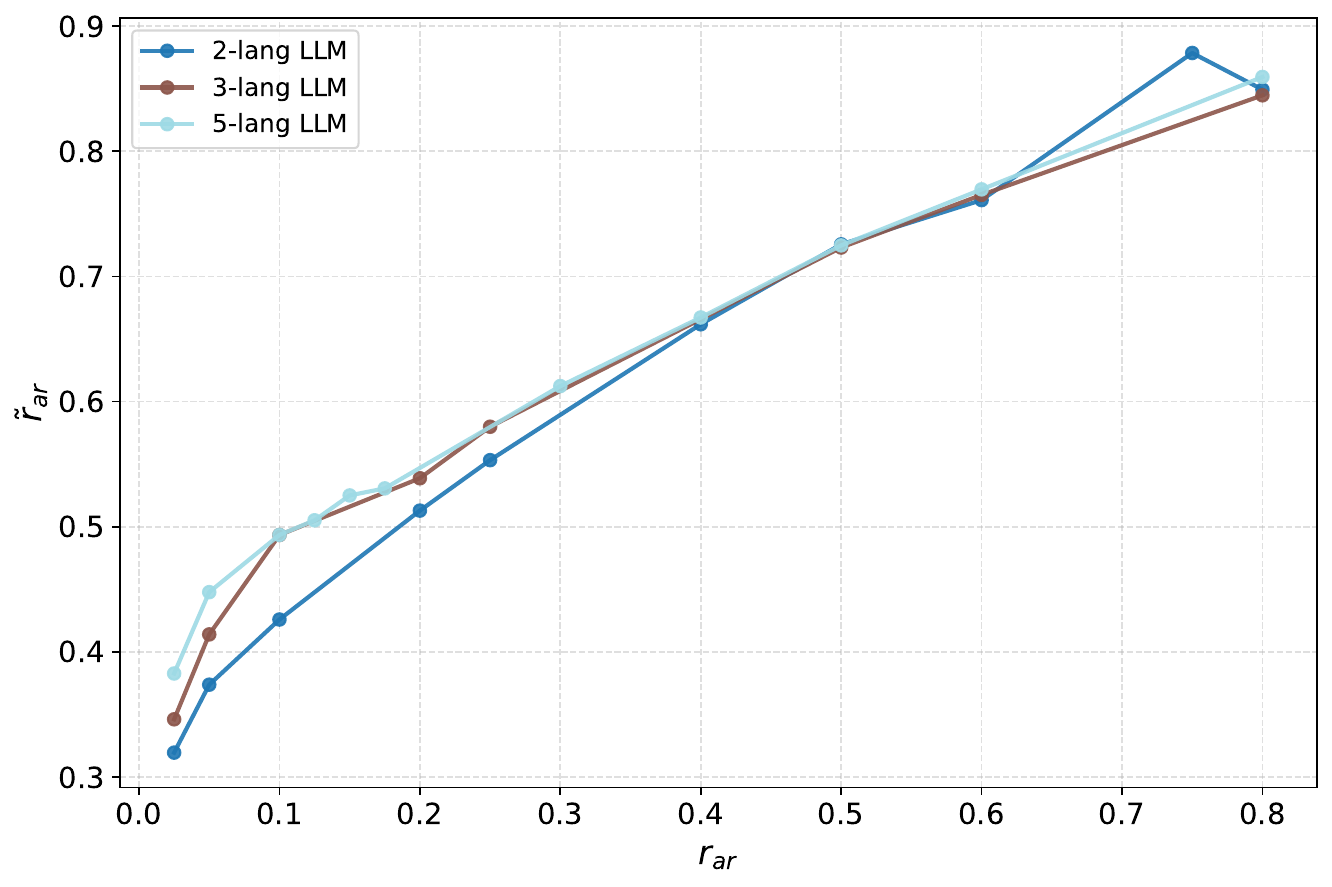}
    }\hfill
    \subfigure[$\tilde{r}_{ar}$ as a function of $r_{ar}$ with varying training data scales]{
        \includegraphics[width=0.3\textwidth]{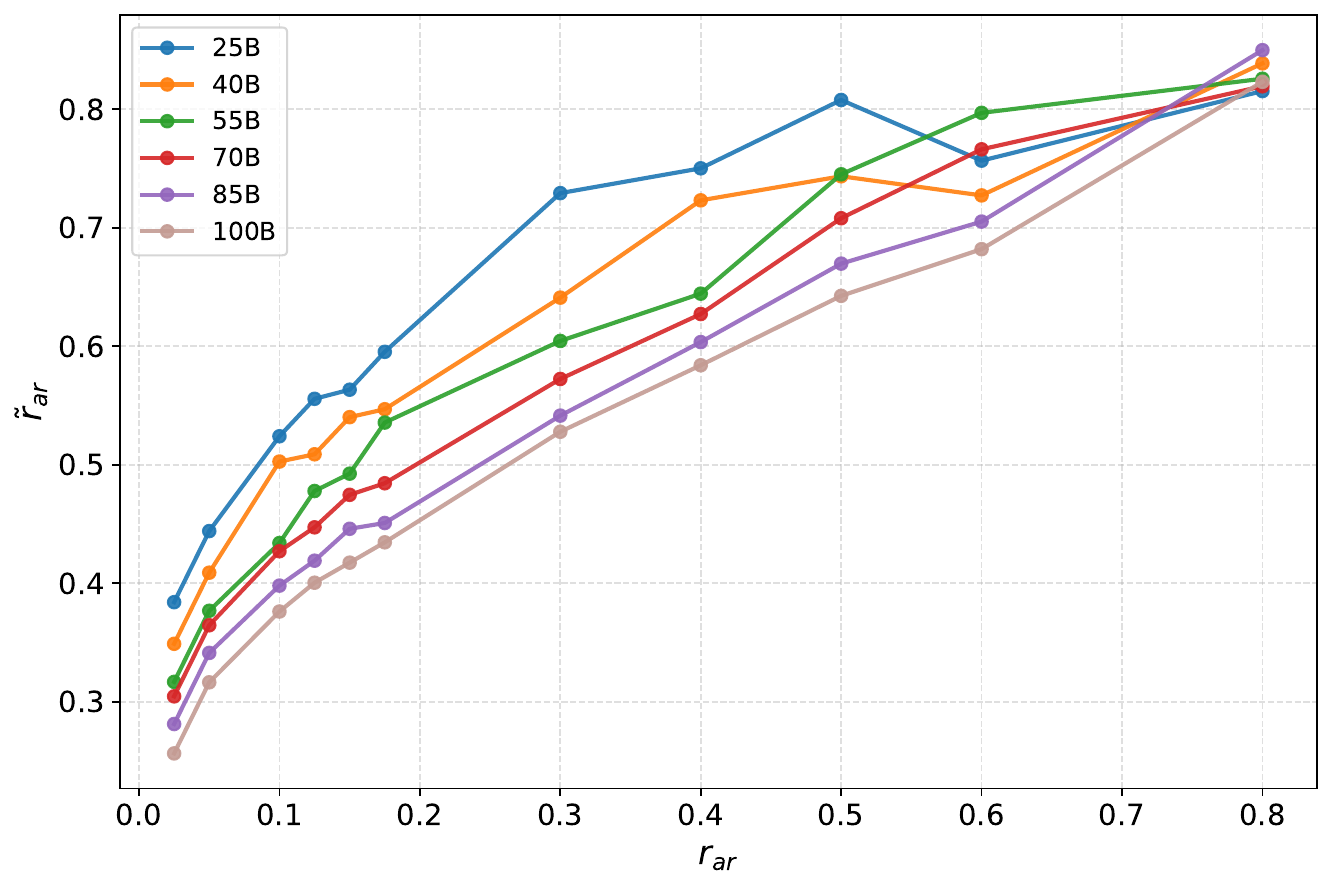}
    }
    \caption{Illustration of cross-lingual interaction-aware language ratio ($\tilde{r}_{ar}$) and its dependency on original training proportions ($r_{ar}$).}
    \label{fig:relation_tilder_r}
\end{figure}

Given a total token budget $D$, each language $L_i$ contributes $D_i = \lfloor r_i \cdot D \rfloor$ tokens to the training set. The model parameters $\theta$ are trained via empirical risk minimization: $\theta^*(D, \mathbf{r}) = \arg\min_{\theta} \mathcal{L}(\theta; D, \mathbf{r})$, where $\mathcal{L}(\theta;D,\mathbf{r})$ denotes the next-token prediction loss on the multilingual training set defined by proportions $\mathbf{r}$ and token budget $D$.

To evaluate the pretrained model, we measure validation loss on a language-specific held-out set $D_i^v$: $\mathcal{L}_i^v(\theta^*(D, \mathbf{r})) = \mathcal{L}(\theta^*(D, \mathbf{r}); D_i^v)$.

Our goal is to identify the optimal language proportion vector $\mathbf{r}^*$ that achieves balanced multilingual performance by minimizing a weighted sum of validation losses across all languages:

\begin{equation}
\footnotesize
\mathbf{r}^* = \arg\min_{\mathbf{r} \in \mathcal{R}^m} \sum_{i=1}^{m} \omega_i \cdot \mathcal{L}_i^v(\theta^*(D, \mathbf{r})),
\end{equation}

where hyperparameter $\omega_i \geq 0$ specify the relative importance of each language $L_i$, set according to application-specific requirements or practical considerations.

This formulation defines a bi-level optimization problem, in which the outer optimization seeks optimal language proportions, and the inner optimization involves training an LLM given these language proportions. Due to the intrinsic complexity of cross-lingual interactions and the prohibitive computational cost of repeated model retraining, directly solving this optimization problem through standard gradient-based approaches is computationally infeasible.

\subsection{Cross-Lingual Interaction-aware Language Ratio}

Given a total token budget \(D\) and a language proportion vector \(\mathbf{r}\), we can obtain the validation loss \(\mathcal{L}_i^v(D, \mathbf{r})\) for language \(L_i\). 
Let \(\tilde{D}_i\) be the number of tokens required to reach the validation loss \(\mathcal{L}_i^v(D, \mathbf{r})\) from a monolingual model solely on language \(L_i\), we then define the \textbf{cross-lingual interaction-aware ratio} \(\tilde{r}_i\) as the ratio of this equivalent monolingual token budget \(\tilde{D}_i\) to the actual multilingual token budget \(D\):
\(
\tilde{r}_i = \frac{\tilde{D}_i}{D}.
\)
Formally, \(\tilde{r}_i\) can be formally expressed as:

\begin{equation}
\footnotesize
\tilde r_i = \frac{1}{D}\left(\frac{B_i}{\mathcal{L}_i^v(D,\mathbf{r}) - E_i}\right)^{1/\beta_i},
\end{equation}
where parameters \(B_i\), \(\beta_i\), and \(E_i\) are derived from the monolingual scaling law \cite{10.5555/3600270.3602446}. Specifically, the monolingual scaling law characterizes how the validation loss decreases as training data volume increases for a single language \(L_i\), expressed as:
\begin{equation}
\footnotesize
\mathcal{L}_i^v(D_i, r_i=1) = \frac{B_i}{D_i^{\beta_i}} + E_i,
\label{eq:basic_scaling_law}
\end{equation}
where \(D_i\) represents the token budget allocated exclusively to language \(L_i\). In the absence of cross-lingual transfer, the interaction-aware ratio \(\tilde r_i\) equals the actual ratio \(r_i\), thus the difference \(\tilde r_i - r_i\) quantifies the magnitude of cross-lingual effects from other languages.

\subsubsection{Empirical Observations and Insights}

To systematically understand the behavior of the cross-lingual interaction-aware language ratio $\tilde r_i$, we conducted over 300 experiments on Transformer-based models. Specifically, we varied the number of jointly trained languages (2, 3, and 5 languages), total token budgets ranging from 5 billion to 100 billion tokens, and explored a wide range of language proportion vectors \(\mathbf{r}\). For each configuration, we computed the pairs \((r_i, \tilde r_i)\) to examine how the effective language ratio deviates from the actual proportion due to cross-lingual transfer. These results are visualized in Figure~\ref{fig:relation_tilder_r}, from which we identify following key empirical insights:





\begin{itemize}

\item \textbf{Dependency on absolute language proportion.} Cross-lingual transfer strength diminishes as the actual language proportion ($r_i$) increases, with the slope gradually decreasing and approaching linearity at higher proportions, as illustrated in Figures~\ref{fig:relation_tilder_r} (a), (b), and (c).

\item \textbf{Dependency on co-training languages.} The specific set of co-training languages affects cross-lingual transfer primarily when the language proportion $r_i$ is small, as demonstrated in Figure~\ref{fig:relation_tilder_r} (a). This influence diminishes as $r_i$ grows.

\item \textbf{Dependency on model language counts.} Increasing the number of co-trained languages affects the intercept rather than the slope of the cross-lingual transfer relationship. This variation shifts the onset point at which transfer strength approaches linearity, as shown in Figure~\ref{fig:relation_tilder_r} (b).

\item \textbf{Dependency on data scale.} Cross-lingual transfer consistently weakens with larger total token budgets ($D$), indicating that increased training data volume reduces dependency between languages, as depicted in Figure~\ref{fig:relation_tilder_r} (c).
\end{itemize}



\subsubsection{Parametric Modeling of Cross-Lingual Interaction-aware Ratio}

Motivated by the empirical insights described above, we propose a parametric model to capture the relationship between the cross-lingual interaction-aware language ratio \(\tilde r_i\) and the actual language ratio \(r_i\). Specifically, we model \(\tilde r_i\) as:
\begin{equation}
\footnotesize
    \tilde r_i = r_i + \left(\sum_{j \neq i}\alpha_{j\rightarrow i}(D)\cdot r_j\right)\left(1 - e^{-\eta_i r_i}\right),
\end{equation}
where the parameters are defined as follows:
\begin{itemize}
    
\item \textbf{\(\alpha_{j\rightarrow i}(D)\)} represents the transfer strength from language \(L_j\) to language \(L_i\). Empirically, we find that this transfer effect diminishes linearly with increasing token budget \(D\), which we model as: $ \alpha_{j\rightarrow i}(D) = b_{ji}+\frac{k_{ji}}{D}$, where \(b_{ji}\) indicates the initial strength of cross-lingual transfer from \(L_j\), and \(k_{ji}\) quantifies the rate at which this transfer strength decays as data volume increases. Details about \(\alpha_{j\rightarrow i}(D)\) is in Appendix A.

\item \(\eta_i\) captures the intrinsic data sufficiency of language \(L_i\). A larger value of \(\eta_i\) indicates that language \(L_i\) remains reliant on cross-lingual transfer across a wider range of proportions, exhibiting a pronounced curved (transfer-dominated) regime. Conversely, a smaller \(\eta_i\) signals that language \(L_i\) quickly enters a linear (self-dominated) regime, reflecting sufficient self-contained data.
\end{itemize}

\subsubsection{Complete Cross-Lingual Interaction-aware Scaling Law}

By incorporating the parametric definition of the cross-lingual interaction-aware ratio into the monolingual scaling law, we obtain our final scaling law formulation:
\begin{align}
\footnotesize
\mathcal{L}_i^v(D,\mathbf{r}) 
&= \frac{B_i}{\left[D \cdot \tilde r_i\right]^{\beta_i}} + E_i \\
&= \frac{B_i}{\left[D \cdot \left(r_i + \left(\sum_{j\neq i}(b_{ji}+\frac{k_{ji}}{D})\cdot r_j\right)\cdot(1 - e^{-\eta_i r_i})\right)\right]^{\beta_i}} + E_i.
\label{eq:final_scaling_law}
\end{align}

In this equation, the complete set of parameters to estimate are: $\{B_i,\beta_i,E_i\}_{i=1}^{m}, \, \{b_{ji},k_{ji}\}_{i, j=1, j\neq i}^{m}, \, \{\eta_i\}_{i=1}^{m}$.

\paragraph{Parameter Estimation Procedure.}
To fully determine these parameters, we perform targeted experiments involving each language individually. Specifically, for each language \(L_i\), we conduct three experiments with distinct proportions: one monolingual scenario (where \(r_i=1\)) to estimate the baseline scaling law parameters \(B_i\), \(\beta_i\), and \(E_i\), and two additional multilingual experiments with randomly chosen language proportions \(r_i\) and the remaining languages allocated equally as \(\frac{1-r_i}{m-1}\). Each of these experiments is repeated at two distinct training token budgets to ensure reliable parameter fitting across data scales. Following the experimental setup \cite{10.5555/3600270.3602446}, we fit our scaling law parameters only using data points from the last 15\% of training. Thus, for a setting with \(m\) languages, this structured approach requires a total of \(3 \times m \times 2\) experiments, enabling comprehensive and accurate estimation of the proposed scaling law parameters. The detailed fitting procedure is summarized in Algorithm~\ref{alg:combined}.

\begin{algorithm}[t]
\caption{\textsc{Climb}}
\label{alg:combined}
\begin{algorithmic}[1]
\Require Languages $\{L_1,\dots,L_m\}$, token budgets $\{D^{(1)}, D^{(2)}\}$.
\Ensure Parameters $\{B_i,\beta_i,E_i\}_{i=1}^{m}, \, \{b_{ji},k_{ji}\}_{i, j=1, j\neq i}^{m}, \, \{\eta_i\}_{i=1}^{m}$, optimal language proportions $\mathbf{r}^*$.

\Statex \textbf{\textit{Part I: Parameter Modeling of Cross-Lingual Interaction-aware Language Ratio}}
\For {each language $L_i$}
    \State Conduct monolingual experiments ($r_i=1$) at $D^{(1)}, D^{(2)}$.
    \State Fit monolingual scaling law \ref{eq:basic_scaling_law} to estimate $B_i$, $\beta_i$, $E_i$.
    \For {each proportion $r_i=c_i \in (0,1)$, repeat twice}
        \State Set other languages proportion $r_j = \frac{1-c_i}{m-1}$, $\forall j \neq i$.
        \For {each token budget $D \in \{D^{(1)}, D^{(2)}\}$}
            \State Train model with proportions $\mathbf{r}$ and budget $D$.
            \State Record validation loss $\mathcal{L}_i^v(D,\mathbf{r})$.
            \State Compute $\tilde r_i$ from Eq. (\ref{eq:final_scaling_law}).
        \EndFor
    \EndFor
    \State Fit parameters $b_{ji}$, $k_{ji}$, $\eta_i$ using $(r_i,\tilde r_i)$ pairs.
\EndFor

\Statex \textbf{\textit{Part II: Estimating Optimal Multilingual Balanced Allocation}}
\State Compute optimal direction components $p_i$ via Eq.~\eqref{eq:optimal_direction}.
\State Normalize direction: $\hat p_i \gets p_i/\sum_j p_j$ for all $i$.
\State Solve constrained optimization (Eq.~\eqref{eq:optimal_magnitude}).

\State \Return parameters $\{B_i,\beta_i,E_i\}_{i=1}^{m}, \, \{b_{ji},k_{ji}\}_{i, j=1, j\neq i}^{m}, \, \{\eta_i\}_{i=1}^{m}$, and optimal proportions $\mathbf{r}^*$.
\end{algorithmic}
\end{algorithm}

\subsection{Estimating Optimal Multilingual Balanced Allocation}

Directly minimizing the multilingual validation loss defined by Equation~\eqref{eq:final_scaling_law} is challenging, as it forms a non-convex optimization problem in language proportions \(\mathbf{r}\). While it may appear intuitive to directly optimize the cross-lingual interaction-aware language ratios \(\tilde{r}_i\) under Equation \eqref{eq:final_scaling_law}, this objective is intractable in practice, as the total sum \(\sum_i \tilde{r}_i\) remains unknown. To address this difficulty, we propose a two-stage optimization procedure that decomposes the original complex problem into two simpler, sequential steps. Specifically, we first determine the optimal direction in the cross-lingual interaction-aware language ratios \(\tilde r_i\) space, ensuring balanced marginal benefits across languages. Subsequently, we optimize the magnitude along this determined direction to identify the final allocation \(\mathbf{r}\) that maximizes the overall cross-lingual interaction-aware language ratios \(\tilde r_i\), effectively minimizing the multilingual validation loss. 

\subsubsection{Optimal Direction via Marginal-Benefit Balancing.}

In the first stage, we identify the optimal direction for the cross-lingual interaction-aware language ratios \(\tilde r_i\) by balancing the marginal benefits across all languages. Specifically, we derive the optimal proportional relationship between the interaction-aware ratios by equalizing the marginal validation-loss reduction contributed by each language. The resulting optimal direction \(p_i\) for each language \(L_i\) is formally given by (see detailed derivation in Appendix B):
\begin{equation}
\footnotesize
p_i = \frac{\left(\omega_i B_i \beta_i\right)^{1/(\beta_i+1)} D^{-\beta_i/(\beta_i+1)}}
{\sum_{k=1}^m \left(\omega_k B_k \beta_k\right)^{1/(\beta_k+1)} D^{-\beta_k/(\beta_k+1)}},
\label{eq:optimal_direction}
\end{equation}
where \(B_i\) and \(\beta_i\) are the monolingual scaling-law parameters of language \(L_i\), and \(\omega_i\) represents the predefined importance weight for language \(L_i\). Intuitively, the direction \(p_i\) indicates the ideal relative allocation of interaction-aware language ratios, balancing each language's data efficiency, validation-loss reduction rate, and relative importance. Identifying this optimal direction substantially reduces complexity in subsequent optimization steps by constraining the search space for the final language proportions.

\subsubsection{Optimal Magnitude via Constrained Effective Allocation Maximization.}

With the optimal direction \(p_i\) identified, the second stage focuses on determining the optimal magnitude along this direction. Nevertheless, due to the monotonicity of the scaling law function \ref{eq:basic_scaling_law}, we find that a larger aggregate \(\sum_i \tilde{r}_i\) consistently implies a lower overall training loss, thereby revealing an implicit preference for maximizing effective data contributions across languages (details in Appendix C). Specifically, we recover the actual language proportions \(\mathbf{r}\) by solving a constrained optimization problem that maximizes the total cross-lingual interaction-aware language ratio while staying close to the previously determined direction \(p\). Formally, this optimization objective is defined as:
\begin{equation}
\footnotesize
\min_{\mathbf{r}} \left[-\sum_{i=1}^m \hat r_i(\mathbf{r}) + \rho \sum_{i=1}^m (\hat r_i(\mathbf{r}) - p_i)^2 \right], \quad\text{s.t.}\quad \sum_{i=1}^m r_i=1,\quad r_i\ge 0,
\label{eq:optimal_magnitude}
\end{equation}
where \(\hat r_i=\frac{\tilde r_i}{\sum_j \tilde r_j}\) is the normalized interaction-aware ratios and direction components, respectively. The first term of the objective function aims at maximizing the overall interaction-aware language ratio, corresponding directly to minimizing multilingual validation loss, while the second term (soft-constraint) penalizes deviations from the optimal direction \(p\). The hyperparameter \(\rho>0\) balances these two objectives, with smaller values of \(\rho\) emphasizing pure loss minimization and larger values enforcing adherence to the optimal direction.

This problem is inherently non-convex due to the interaction-aware ratio's nonlinearity. Thus, we adopt a Trust-Region Interior-Point Method to efficiently handle this constrained optimization problem. This structured reformulation significantly reduces the complexity and dimensionality of the original allocation problem, accelerating convergence and improving numerical stability.

\section{Crosslingual Interaction-aware Language Ratio Evaluation}

\subsection{Experimental Setup}

\paragraph{Model Architecture.}  
We utilize the LLaMA-2 \citep{touvron2023llama2openfoundation} architecture with 1.2 billion parameters, training all models from scratch with randomly initialized weights. All experiments are condcuted on Nvidia H100 GPU cards. To ensure consistency with established scaling-law practices, we follow the Chinchilla configuration and set a fixed number of training steps for each dataset size, adjusting the learning-rate decay schedule to cosine. Validation losses are calculated by averaging the results obtained from the final three training steps. 

\paragraph{Datasets.}  
All experiments are conducted using data sampled from the Fineweb-2 corpus \cite{penedo2024fineweb-2}. To rigorously evaluate our Cross-Lingual Interaction-aware Language Ratio across diverse linguistic scenarios, we conduct experiments involving models trained on 2, 3, 5, and 16 languages, respectively. For each multilingual setting, we vary the token budgets from 5 billion to 100 billion tokens. The specific language compositions and detailed training procedures are documented in the Appendix D.

\begin{table}[t]
\centering
\setlength{\tabcolsep}{12pt}
\caption{Fitting and Extrapolation Performance of Different Methods ($R^2\!\uparrow$ and Huber Loss$\downarrow$) for Multilingual LLMs at 100B and 1T Tokens.}
\resizebox{\textwidth}{!}{
\begin{tabular}{@{}lcccccccc@{}}
\toprule
& \multicolumn{2}{c}{2-lang LLM} & \multicolumn{2}{c}{3-lang LLM} & \multicolumn{2}{c}{5-lang LLM} & \multicolumn{2}{c}{16-lang LLM} \\ 
\cmidrule(lr){2-3} \cmidrule(lr){4-5} \cmidrule(lr){6-7} \cmidrule(lr){8-9}
& $R^2$\,$\uparrow$ & \makecell{Huber\,$\downarrow$\\($\times 10^{-3}$)} 
& $R^2$\,$\uparrow$ & \makecell{Huber\,$\downarrow$\\($\times 10^{-3}$)} 
& $R^2$\,$\uparrow$ & \makecell{Huber\,$\downarrow$\\($\times 10^{-3}$)} 
& $R^2$\,$\uparrow$ & \makecell{Huber\,$\downarrow$\\($\times 10^{-3}$)} \\ 
\midrule
\multicolumn{9}{l}{\textbf{Fitting Results (Total Training Tokens: 100B)}}\\ 
\midrule
Isolated & 0.649 & 7.95 & 0.743 & 5.35 & 0.734 & 5.34 & 0.768 & 5.26\\
MSL & 0.832 & 5.61 & 0.854 & 2.15 & 0.823 & 1.94 & 0.836 & 2.20\\
\textsc{Climb} & \textbf{0.978} & \textbf{0.518} & \textbf{0.986} & \textbf{0.301} & \textbf{0.992} & \textbf{0.205} & \textbf{0.981} & \textbf{0.274}\\ 
\midrule
\multicolumn{9}{l}{\textbf{Extrapolation Results (Total Training Tokens: 1T)}}\\ 
\midrule
Isolated & 0.648 & 8.21 & 0.741 & 5.38 & 0.732 & 5.36 & 0.767 & 5.30\\
MSL & 0.830 & 5.79 & 0.852 & 2.24 & 0.822 & 1.98 & 0.834 & 2.24\\
\textsc{Climb} & \textbf{0.964} & \textbf{0.525} & \textbf{0.947} & \textbf{0.310} & \textbf{0.948} & \textbf{0.208} & \textbf{0.936} & \textbf{0.278}\\ 
\bottomrule
\end{tabular}
}
\label{tab:performance_llms}
\end{table}

\paragraph{Evaluation Metrics \& Baseline.}  
We assess the accuracy of our validation-loss predictions primarily using two metrics: the coefficient of determination ($R^2$) and the Huber loss. The $R^2$ score measures the proportion of variance explained by our fitted scaling-law model, with values closer to 1 indicating greater predictive accuracy. Additionally, we employ Huber loss, a robust error metric combining properties of mean squared error and mean absolute error, which provides resilience against outliers; lower Huber loss values reflect more accurate predictions. 

\paragraph{Baselines}
We compare our approach against two baselines: 1) an assumption of no cross-lingual transfer, where each language's validation loss depends solely on its own proportion, labeled as ``isolated''; and 2) a recent multilingual scaling-law study~\citep{he2024scalinglawsmultilinguallanguage}, referred to as ``MSL''.

\begin{figure}[t]
    \centering
    \begin{minipage}[b]{0.32\textwidth}
        \centering
        \includegraphics[width=\textwidth]{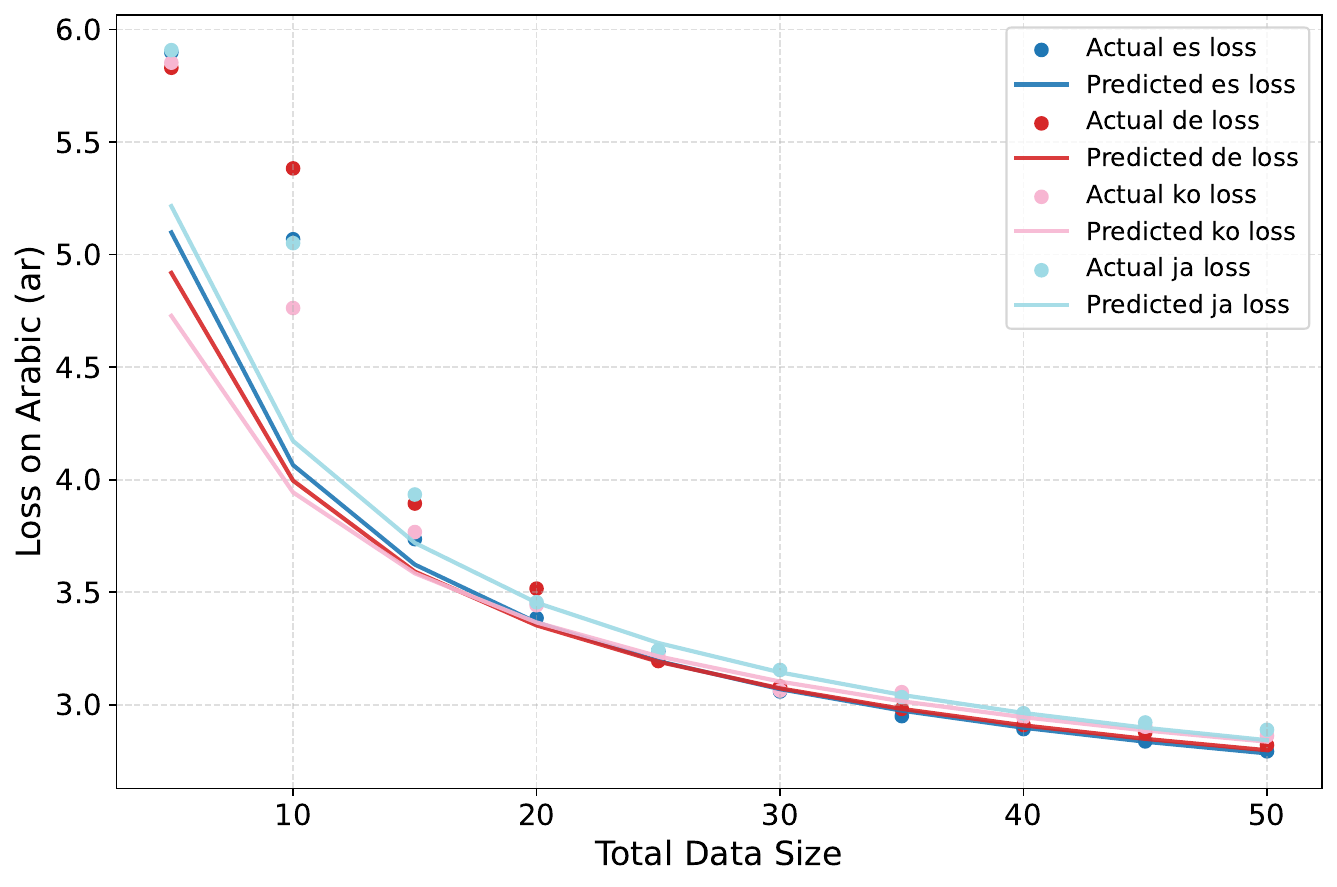}
        \caption{Curve fitting on Arabic data (5-lang LLM). Solid line: fitted results.}
        \label{fig:ar_fit}
    \end{minipage}\hfill
    \begin{minipage}[b]{0.32\textwidth}
        \centering
        \includegraphics[width=\textwidth]{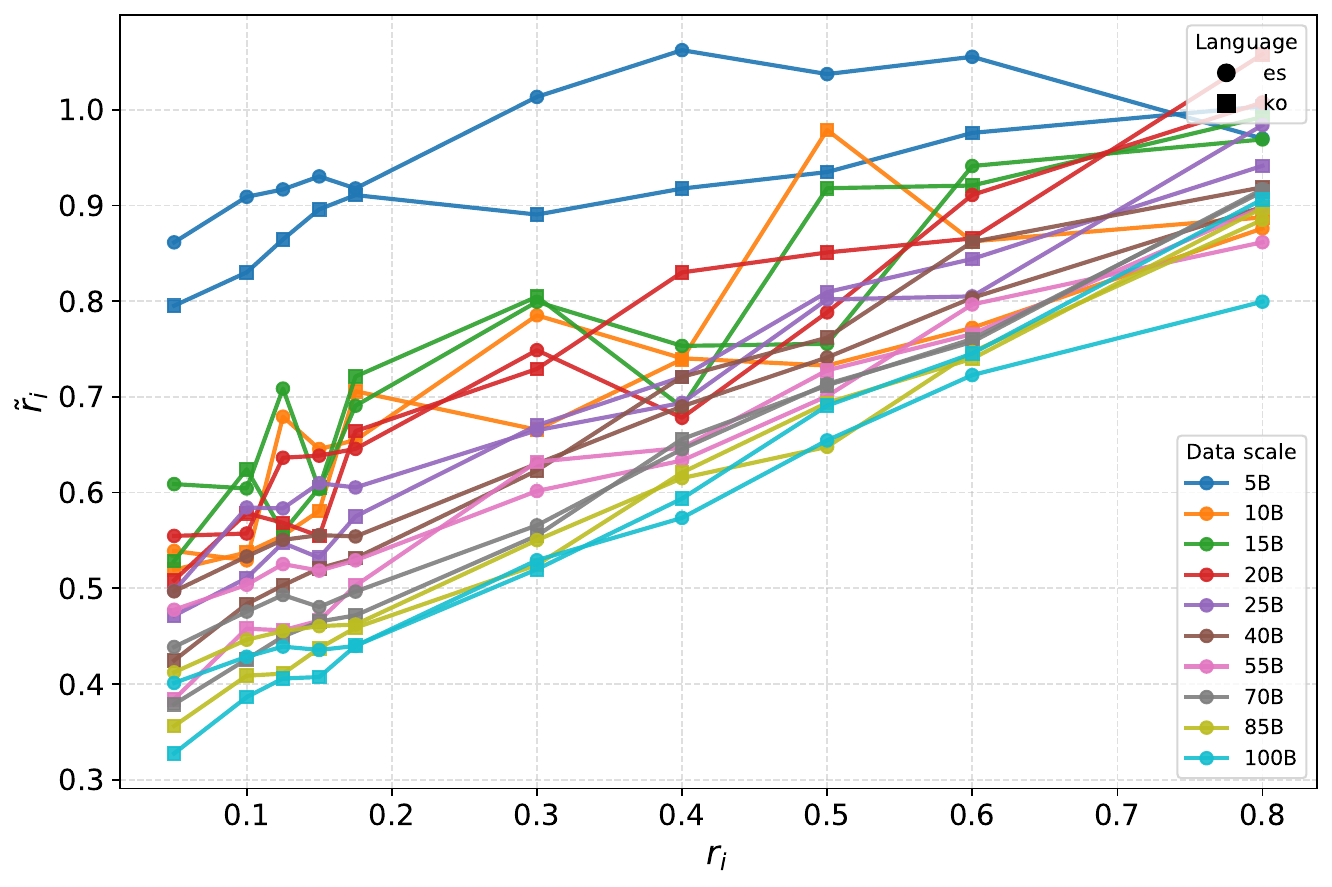}
        \caption{$\tilde{r}$ vs. $r$ across corpus scales on 2 different languages (es, ko).}
        \label{fig:prove_r_tilde_r}
    \end{minipage}\hfill
    \begin{minipage}[b]{0.32\textwidth}
        \centering
        \includegraphics[width=\textwidth]{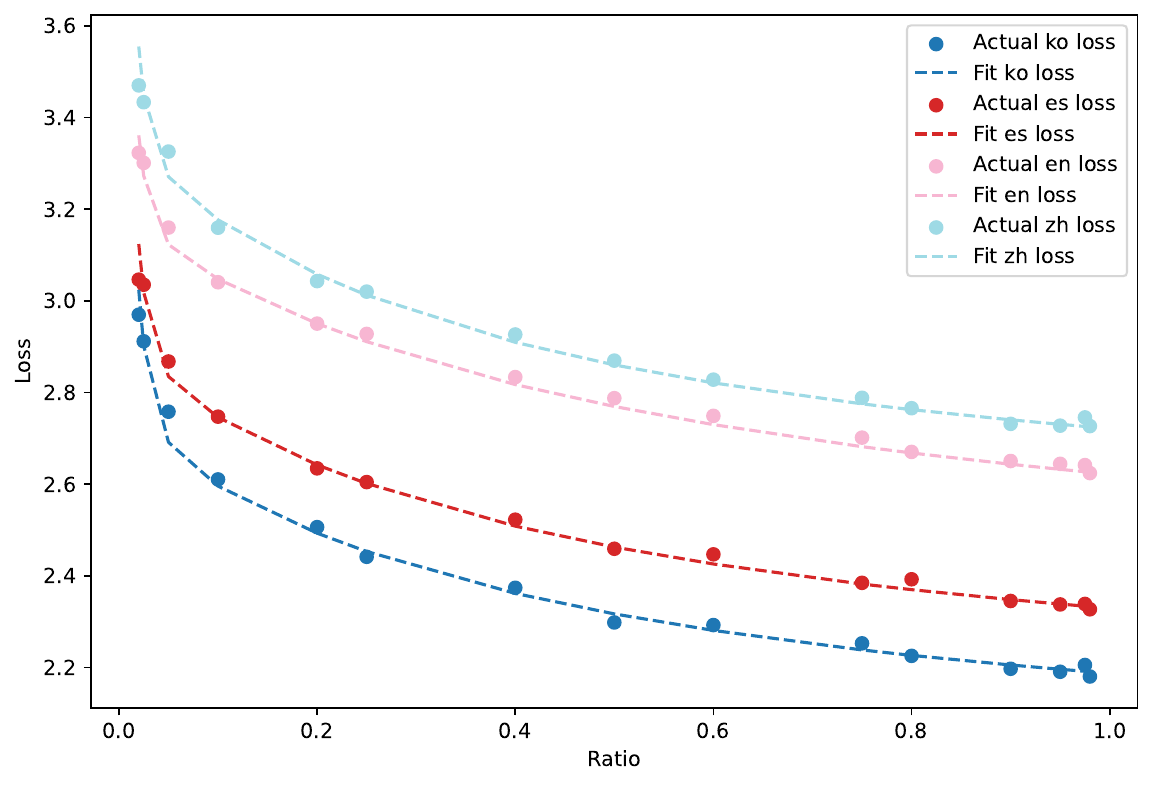}
        \caption{Validation loss fitting across language ratios (0.01–0.99).}
        \label{fig:fitting_r_capability}
    \end{minipage}
\end{figure}

\subsection{Scaling Law Fit Accuracy}

We present the prediction errors of our proposed scaling law compared to the baseline (MSL) in Table~\ref{tab:performance_llms}, evaluating models trained across various multilingual scenarios (2–16 lang LLMs) with token budgets of 100\,B and 1\,T tokens. Our cross-lingual interaction-aware approach consistently achieves lower prediction errors compared to the baseline, effectively capturing validation-loss trends in both homogeneous (same language-family) and heterogeneous language settings. From 2-lang LLM to 16-lang LLM, \textsc{Climb}'s Huber loss remains consistently an order of magnitude lower than both baselines, highlighting the importance and prevalence of cross-lingual transfer effects in multilingual models.
Conversely, our scaling-law formulation remains robust and delivers accurate loss predictions even in highly complex multilingual scenarios.

\subsection{Scaling Law Applicability}

To comprehensively evaluate the applicability and robustness of our scaling law, we explore its validity across different training scales and language proportions (see Figure~\ref{fig:ar_fit} and Figure~\ref{fig:prove_r_tilde_r}). At smaller training scales (below 25\,B tokens), our model's predictive accuracy is limited due to unstable cross-lingual transfer relationships in low-data regimes. However, prediction accuracy markedly improves for data budgets exceeding 25\,B tokens, as the cross-lingual interactions become stable and predictable. Additionally, we extrapolate predictions up to 1\,T tokens, observing excellent alignment between our predicted curves and the empirical validation-loss data, as shown in Table~\ref{tab:performance_llms}. Further experiments from Figure~\ref{fig:fitting_r_capability} in bilingual settings demonstrate our formula's capability to precisely fit the validation loss across the entire feasible range of language proportions (from 0.01 to 0.99), underscoring the broad applicability of our scaling law in practical multilingual model deployments.

\section{Multilingual Balanced Allocation Performance}

\subsection{Experimental Setup}

\paragraph{Model Architecture and Training Setup.}
We evaluate multilingual performance by training Transformer models based on the LLaMA-2 \citep{touvron2023llama2openfoundation} architecture at two different scales: 1.2\,B and 7\,B parameters. All models are trained using the Fineweb-2 corpus, identical to the datasets employed in scaling-law experiments, with each model ingesting a total of 1\,T tokens. Peak learning rates are set to $4.3 \times 10^{-5}$ for the 1.2\,B model and $3.6 \times 10^{-5}$ for the 7\,B model, both following a cosine-decay schedule that decays the learning rate down to 10\% of its initial value.

\paragraph{Baselines.}
We compare our proposed \textsc{Climb}-derived allocations against two categories of baselines. First, we evaluate against state-of-the-art publicly available multilingual models, specifically, LLaMA-3.2-1B \citep{grattafiori2024llama3herdmodels}, GEMMA-3-1B-pt \citep{gemmateam2025gemma3technicalreport} and Qwen-3-1.7B-base \citep{yang2025qwen3technicalreport}, which are either trained on significantly larger datasets or distilled from larger models. Second, under identical model architecture and data volume constraints, we train models using three alternative language allocation strategies: (1) uniform allocation across all 16 languages; (2) allocations derived independently from individual monolingual scaling laws; and (3) allocations derived from the existing MSL formula, which assumes language-family independence. (4) natural proportions reflecting the frequency of each language's original dataset.

\paragraph{Evaluation Benchmarks.}
To comprehensively evaluate \textsc{Climb}, we translate several English benchmarks into multilingual to further assess model performance; benchmarks translated by us are marked \textsuperscript{\ddag} in Table~\ref{tab:multilingual_evaluation}. Specifically, we adopt the following tasks: language modeling (\textit{XNLI}~\citep{conneau-etal-2018-xnli}, \textit{XCOPA}~\citep{ponti-etal-2020-xcopa}, \textit{XStoryCloze}(XSC)~\citep{lin-etal-2022-shot}), commonsense reasoning (\textit{HellaSwag}(XHS)~\citep{zellers-etal-2019-hellaswag}, \textit{XWinograd}(XWG)~\citep{muennighoff-etal-2023-crosslingual}), question answering (\textit{ARC-Easy/ARC-Challenge}~\citep{clark2018thinksolvedquestionanswering}, \textit{GPQA} \citep{
rein2024gpqa}, \textit{TRUTHFULQA}(TQA) \citep{lin-etal-2022-truthfulqa}), MMLU series (\textit{CMMLU} \citep{li-etal-2024-cmmlu}, \textit{JMMLU} \footnote{\url{https://github.com/nlp-waseda/JMMLU}}, \textit{VMLU} \footnote{\url{https://vmlu.ai/}}, \textit{GMMLU} \citep{singh2025globalmmluunderstandingaddressing}), and translation (\textit{FLORES} \citep{nllbteam2022languageleftbehindscaling}). Detailed evaluation protocols for each benchmark are provided in the Appendix E.

\begin{table}[t]
\centering
\caption{Performance of \textsc{Climb} on downstream benchmarks. Results marked with \textsuperscript{\ddag} indicate datasets translated by us. Results for publicly available models reported are re-implemented by us.}
\resizebox{\textwidth}{!}{%
\begin{tabular}{@{}lcccccccccccccc@{}}
\toprule
 & \multicolumn{3}{c}{\textbf{Language Modeling}} & \textbf{Translation} & \multicolumn{4}{c}{\textbf{MMLU}} & \multicolumn{2}{c}{\textbf{Commonsense}} & \multicolumn{4}{c}{\textbf{Question Answering}} \\
\cmidrule(lr){2-4} \cmidrule(lr){5-5} \cmidrule(lr){6-9} \cmidrule(lr){10-11} \cmidrule(lr){12-15}
 & XNLI & XCOPA & XSC & Flores & GMMLU & CMMLU & JMMLU & VMLU & XWG & XHS\textsuperscript{\ddag} & XARC-E\textsuperscript{\ddag} & XARC-C\textsuperscript{\ddag} & XGPQA\textsuperscript{\ddag} & XTQA\textsuperscript{\ddag} \\
\midrule
\multicolumn{15}{l}{\textbf{Open Source Multilingual LLM}} \\
\midrule
LLaMA-3.2 & 40.35 & 57.13 & 58.17 & 44.14 & 28.72 & 30.46 & 29.66 & 28.88 & 75.84 & 42.65 & 47.34 & 31.17 & 24.16 & 37.73 \\
Qwen-3 & 41.33 & 59.55 & 60.51 & 50.15 & 34.28 & 45.17 & 36.09 & 36.15 & 79.16 & 45.42 & 58.26 & 39.31 & 30.83 & 48.88 \\
Gemma-3 & 41.37 & 55.65 & 54.84 & 36.79 & 27.16 & 28.60 & 28.38 & 30.48 & 66.94 & 40.52 & 49.30 & 29.09 & 23.99 & 40.07 \\
\midrule
\multicolumn{15}{l}{\textbf{Different Data Allocation Methods}} \\
\midrule
Uniform & 40.08 & 59.49 & 58.99 & 47.51 & 29.05 & \textbf{34.79} & 32.47 & 31.44 & 73.34 & 48.12 & 59.76 & 35.41 & 26.18 & 39.62 \\
Isolated & 38.93 & 59.42 & 58.74 & 47.58 & 28.64 & 33.78 & 31.85 & 30.91 & 72.31 & 48.11 & 58.53 & 34.78 & 24.58 & 39.71 \\
Natural & 39.05 & 56.54 & 57.38 & 48.54 & 30.23 & 32.10 & 30.94 & 31.12 & 74.98 & 45.68 & 56.32 & 33.54 & 26.37 & 40.63 \\
MSL & 38.54 & 58.06 & 57.94 & 47.75 & 29.00 & 33.24 & 31.49 & 30.25 & 73.09 & 47.02 & 57.60 & 34.17 & 25.36 & 39.49 \\
\textsc{Climb} & \textbf{41.65} & \textbf{59.98} & \textbf{60.54} & \textbf{50.43} & \textbf{31.78} & 33.67 & \textbf{33.21} & \textbf{31.76} & \textbf{77.48} & \textbf{48.75} & \textbf{60.45} & \textbf{36.56} & \textbf{27.03} & \textbf{40.94} \\
\bottomrule
\end{tabular}}
\label{tab:multilingual_evaluation}
\end{table}

\subsection{Results}

Table~\ref{tab:multilingual_evaluation} summarizes the main experimental results, with detailed per-language performance provided in Appendix F. Despite being trained on a relatively modest 1\,T-token budget, our 1.2\,B-parameter model achieves competitive performance against publicly available models such as LLaMA-3.2, Gemma-3, and Qwen-3. When compared directly to different allocation methods, our \textsc{Climb}-derived allocation consistently delivers state-of-the-art results. Notably, we observe absolute performance improvements of up to 2.60\% on the XNLI task compared to isolated allocation. On average, \textsc{Climb} outperforms the baselines by approximately 1.85\% across all evaluated tasks, underscoring the efficacy of our proposed multilingual allocation strategy.


\begin{figure}[t]
    \centering
    \begin{minipage}[b]{0.32\textwidth}
        \centering
        \includegraphics[width=\textwidth]{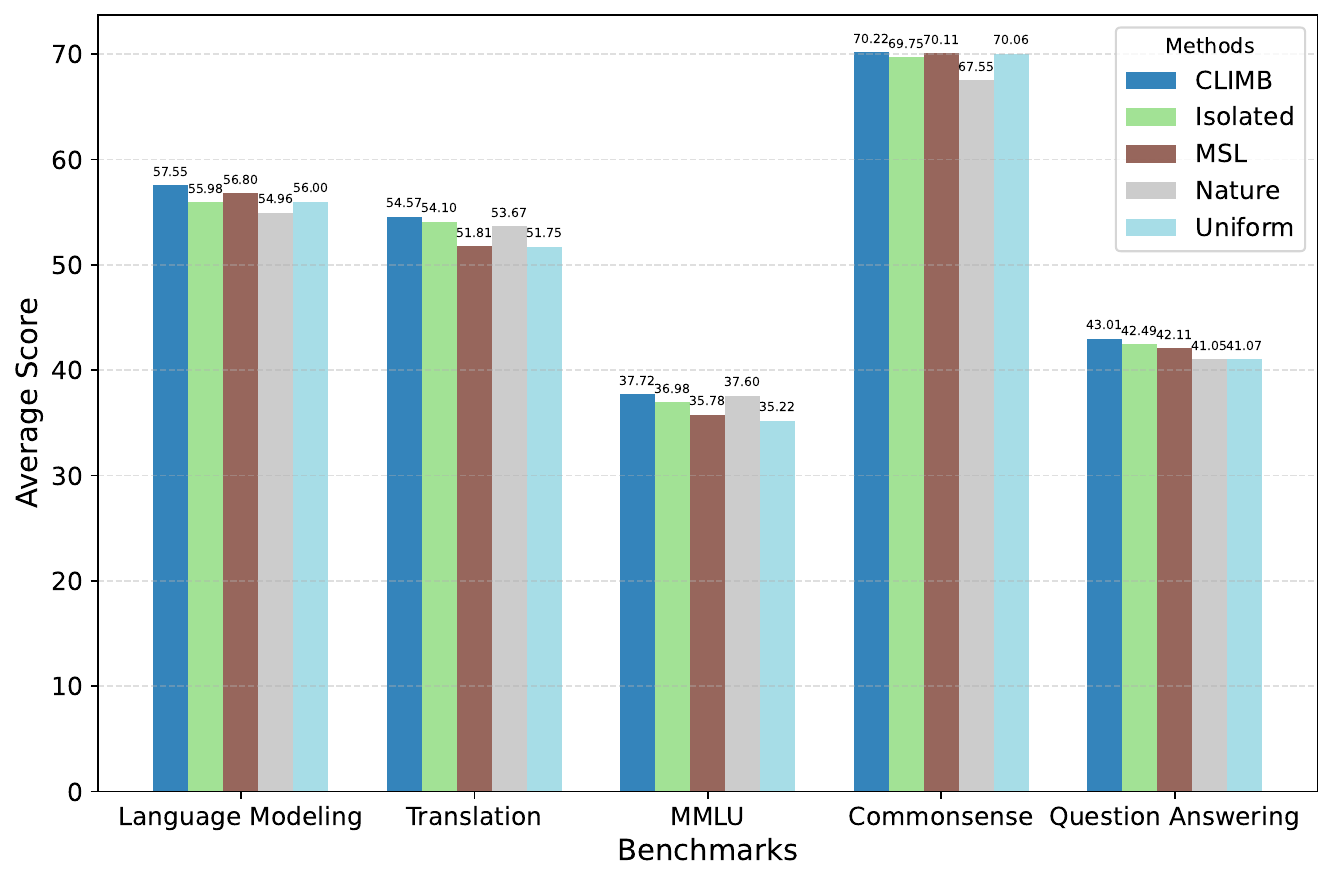}
        \caption{Different allocation results on 7B model.}
        \label{fig:7b_res}
    \end{minipage}\hfill
    \begin{minipage}[b]{0.32\textwidth}
        \centering
        \includegraphics[width=0.7\textwidth]{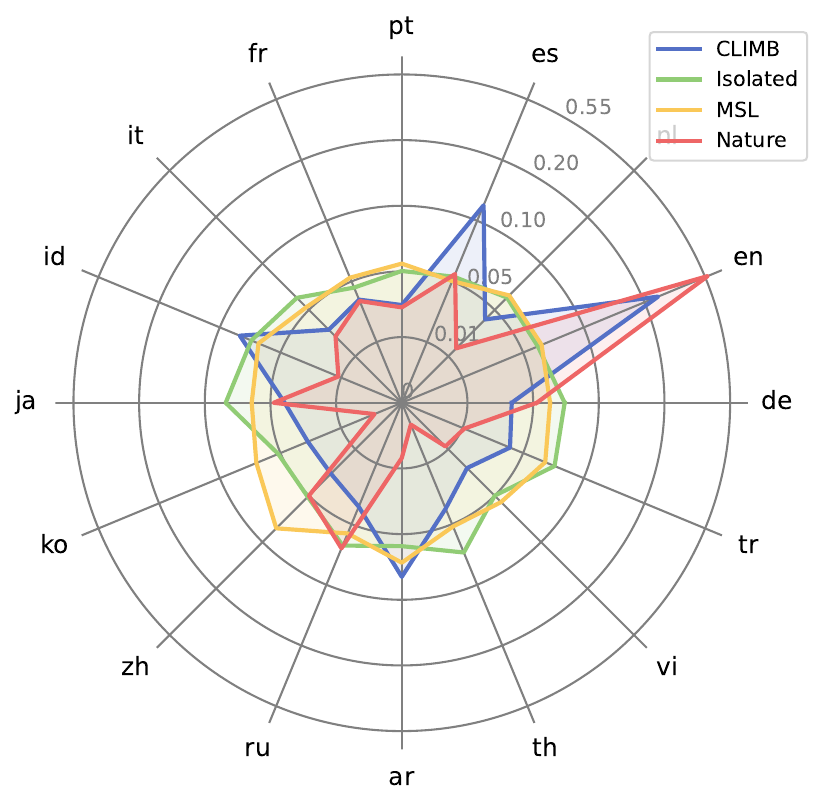}
        \caption{Illustration of each language allocation methods.}
        \label{fig:radar}
    \end{minipage}\hfill
    \begin{minipage}[b]{0.32\textwidth}
        \centering
        \includegraphics[width=\textwidth]{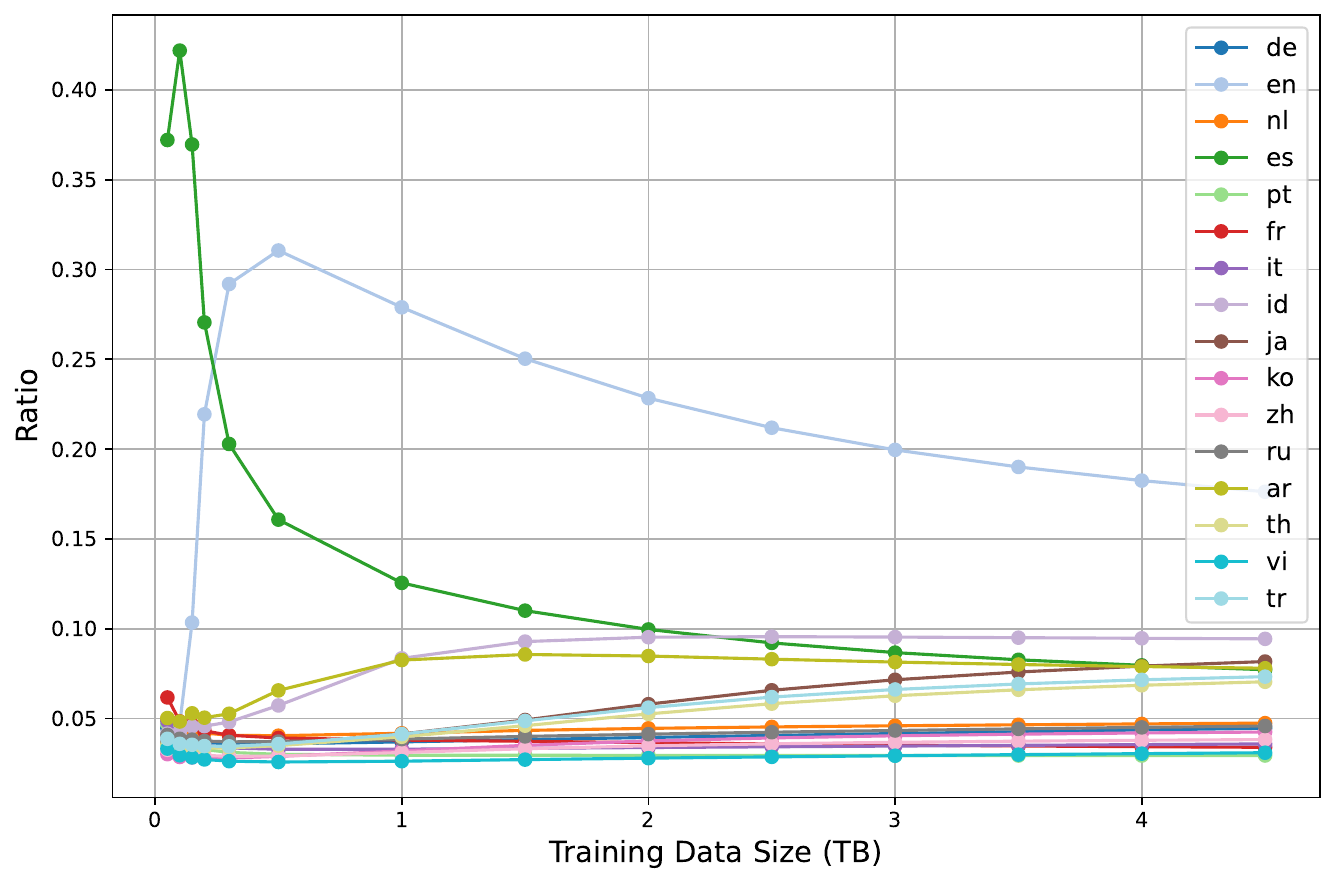}
        \caption{Illustration of how ratios varying with training data.}
        \label{fig:r_change_with_D}
    \end{minipage}
\end{figure}

\paragraph{Generalization to Larger Models.}
Though our optimal language proportions were initially derived using a 1.2B-parameter model, the methodology generalizes effectively to larger scales. We trained a 7B-parameter model using the same total token budget (1T tokens) and evaluated performance (Figure~\ref{fig:7b_res}). \textsc{Climb}-derived allocations consistently outperform baselines by an average of 3.4\%, highlighting improvements across all evaluation dimensions.

\paragraph{Allocation Differences Across Methods.}
To illustrate differences among allocation strategies, Figure~\ref{fig:radar} compares language proportions of various methods. \textsc{Climb} distinctly allocates higher proportions to languages benefiting most from cross-lingual interactions, unlike baselines (\textit{Isolated}, \textit{MSL}, \textit{Nature}), which either allocate evenly or based solely on single-language characteristics. This focused allocation emphasizes the practical advantage of modeling cross-lingual transfer explicitly.

\paragraph{Optimal Language Allocation Shifts with Data Scale.}
Figure~\ref{fig:r_change_with_D} presents how optimal language allocations evolve with increasing token budgets. At smaller scales, simpler or less-resourced languages initially receive higher allocations, quickly lowering validation loss. As data scale increases, allocations shift towards linguistically complex and diverse languages due to their sustained effectiveness at reducing loss. This dynamic trend highlights the necessity of adjusting language proportions according to cross-lingual transfer effects at varying training scales.

\section{Related Work}

\textbf{Data Allocation in Language Model Pretraining.}
Recent studies have explored data allocation strategies to improve LLM pretraining efficiency and downstream performance of LLM pre-training, typically by optimizing weights at the domain \citep{kang2025autoscalescaleawaredatamixing,10.5555/3692070.3692585}, point level \citep{NEURIPS2024_ed165f2f,NEURIPS2024_c4bec0d2}, even token level \citep{NEURIPS2024_3322a9a7,he-etal-2024-softdedup} with respect to validation loss. Early works adopt GroupDRO-based reweighting objectives (e.g., DoReMi~\citep{NEURIPS2023_dcba6be9}) to prioritize difficult domains, while more recent approaches employ influence functions, surrogate models \citep{liu2025regmix,ye2025data}, gradient approximations \citep{thakkar-etal-2023-self,zhang2025harnessing}, or loss-guided heuristics \citep{zhu2025toremitopicawaredatareweighting,sow2025dynamic} to refine data mixture weights under limited training budgets. However, these methods are designed for monolingual English settings, leaving multilingual data allocation underexplored.

\textbf{Scaling Laws for Multilingual LMs.}
Scaling laws~\citep{kaplan2020scalinglawsneurallanguage, henighan2020scalinglawsautoregressivegenerative,10.5555/3600270.3602446,pmlr-v235-ludziejewski24a,qin2025scalinglawssyntheticdata} have become a key tool for predicting language model performance under varying model and data scales, offering practical guidance for efficient resource allocation. While originally developed for monolingual LMs, recent work has recognized the importance of multilingual scaling. Several studies extend scaling analysis to multilingual settings, primarily in neural machine translation~\citep{gordon-etal-2021-data, ghorbani2022scaling, 10.5555/3618408.3618811, zhuocheng-etal-2023-scaling, 10.5555/3666122.3667563,caillaut-etal-2024-scaling,chen2025owlsscalinglawsmultilingual}, yet often under simplified bilingual assumptions and encoder-decoder architectures. Crucially, these works largely ignore cross-lingual transfer effects, which we explicitly model to capture the intricate interactions between languages in multilingual pretraining.

\section{Conclusion}
This paper presents \textsc{Climb}, a novel multilingual optimization framework that effectively models cross-lingual interactions to predict optimal language allocations within a scaling-law paradigm. Empirical evaluations demonstrate that \textsc{Climb} achieves superior predictive accuracy and consistently outperforms baseline methods on multilingual benchmarks at both 1.2B and 7B scales. Nevertheless, our approach currently considers only languages explicitly present in the training data; extending this predictive capacity to languages absent from the training corpus constitutes an important direction for future investigation.

\bibliographystyle{plain}
\bibliography{neurips_2025}

\begin{thebibliography}{10}

\bibitem{caillaut-etal-2024-scaling}
Ga{\"e}tan Caillaut, Mariam Nakhl{\'e}, Raheel Qader, Jingshu Liu, and Jean-Gabriel Barth{\'e}lemy.
\newblock Scaling laws of decoder-only models on the multilingual machine translation task.
\newblock In Barry Haddow, Tom Kocmi, Philipp Koehn, and Christof Monz, editors, {\em Proceedings of the Ninth Conference on Machine Translation}, pages 1318--1331, Miami, Florida, USA, November 2024. Association for Computational Linguistics.

\bibitem{10.5555/3666122.3667563}
Liang Chen, Shuming Ma, Dongdong Zhang, Furu Wei, and Baobao Chang.
\newblock On the pareto front of multilingual neural machine translation.
\newblock In {\em Proceedings of the 37th International Conference on Neural Information Processing Systems}, NIPS '23, Red Hook, NY, USA, 2023. Curran Associates Inc.

\bibitem{chen2025owlsscalinglawsmultilingual}
William Chen, Jinchuan Tian, Yifan Peng, Brian Yan, Chao-Han~Huck Yang, and Shinji Watanabe.
\newblock Owls: Scaling laws for multilingual speech recognition and translation models, 2025.

\bibitem{choenni-etal-2023-languages}
Rochelle Choenni, Dan Garrette, and Ekaterina Shutova.
\newblock How do languages influence each other? studying cross-lingual data sharing during {LM} fine-tuning.
\newblock In Houda Bouamor, Juan Pino, and Kalika Bali, editors, {\em Proceedings of the 2023 Conference on Empirical Methods in Natural Language Processing}, pages 13244--13257, Singapore, December 2023. Association for Computational Linguistics.

\bibitem{clark2018thinksolvedquestionanswering}
Peter Clark, Isaac Cowhey, Oren Etzioni, Tushar Khot, Ashish Sabharwal, Carissa Schoenick, and Oyvind Tafjord.
\newblock Think you have solved question answering? try arc, the ai2 reasoning challenge, 2018.

\bibitem{conneau-etal-2018-xnli}
Alexis Conneau, Ruty Rinott, Guillaume Lample, Adina Williams, Samuel Bowman, Holger Schwenk, and Veselin Stoyanov.
\newblock {XNLI}: Evaluating cross-lingual sentence representations.
\newblock In Ellen Riloff, David Chiang, Julia Hockenmaier, and Jun{'}ichi Tsujii, editors, {\em Proceedings of the 2018 Conference on Empirical Methods in Natural Language Processing}, pages 2475--2485, Brussels, Belgium, October-November 2018. Association for Computational Linguistics.

\bibitem{dam2024completesurveyllmbasedai}
Sumit~Kumar Dam, Choong~Seon Hong, Yu~Qiao, and Chaoning Zhang.
\newblock A complete survey on llm-based ai chatbots, 2024.

\bibitem{deepseekai2025deepseekr1incentivizingreasoningcapability}
DeepSeek-AI.
\newblock Deepseek-r1: Incentivizing reasoning capability in llms via reinforcement learning, 2025.

\bibitem{deepseekai2025deepseekv3technicalreport}
DeepSeek-AI.
\newblock Deepseek-v3 technical report, 2025.

\bibitem{rae2022scalinglanguagemodelsmethods}
Jack W.~Rae et~al.
\newblock Scaling language models: Methods, analysis \& insights from training gopher, 2022.

\bibitem{faisal-anastasopoulos-2024-efficient}
Fahim Faisal and Antonios Anastasopoulos.
\newblock An efficient approach for studying cross-lingual transfer in multilingual language models.
\newblock In Jonne S{\"a}lev{\"a} and Abraham Owodunni, editors, {\em Proceedings of the Fourth Workshop on Multilingual Representation Learning (MRL 2024)}, pages 45--92, Miami, Florida, USA, November 2024. Association for Computational Linguistics.

\bibitem{10.5555/3692070.3692585}
Simin Fan, Matteo Pagliardini, and Martin Jaggi.
\newblock Doge: domain reweighting with generalization estimation.
\newblock In {\em Proceedings of the 41st International Conference on Machine Learning}, ICML'24. JMLR.org, 2024.

\bibitem{10.5555/3618408.3618811}
Patrick Fernandes, Behrooz Ghorbani, Xavier Garcia, Markus Freitag, and Orhan Firat.
\newblock Scaling laws for multilingual neural machine translation.
\newblock In {\em Proceedings of the 40th International Conference on Machine Learning}, ICML'23. JMLR.org, 2023.

\bibitem{gemmateam2024gemma2improvingopen}
Team Gemma.
\newblock Gemma 2: Improving open language models at a practical size, 2024.

\bibitem{gemmateam2024gemmaopenmodelsbased}
Team Gemma.
\newblock Gemma: Open models based on gemini research and technology, 2024.

\bibitem{gemmateam2025gemma3technicalreport}
Team Gemma.
\newblock Gemma 3 technical report, 2025.

\bibitem{ghorbani2022scaling}
Behrooz Ghorbani, Orhan Firat, Markus Freitag, Ankur Bapna, Maxim Krikun, Xavier Garcia, Ciprian Chelba, and Colin Cherry.
\newblock Scaling laws for neural machine translation.
\newblock In {\em International Conference on Learning Representations}, 2022.

\bibitem{gordon-etal-2021-data}
Mitchell~A Gordon, Kevin Duh, and Jared Kaplan.
\newblock Data and parameter scaling laws for neural machine translation.
\newblock In Marie-Francine Moens, Xuanjing Huang, Lucia Specia, and Scott Wen-tau Yih, editors, {\em Proceedings of the 2021 Conference on Empirical Methods in Natural Language Processing}, pages 5915--5922, Online and Punta Cana, Dominican Republic, November 2021. Association for Computational Linguistics.

\bibitem{goyal2024the}
Sachin Goyal, Pratyush Maini, Zachary~Chase Lipton, Aditi Raghunathan, and J~Zico Kolter.
\newblock The science of data filtering: Data curation cannot be compute agnostic.
\newblock In {\em ICLR 2024 Workshop on Navigating and Addressing Data Problems for Foundation Models}, 2024.

\bibitem{grattafiori2024llama3herdmodels}
Aaron Grattafiori and Abhimanyu~Dubey et~al.
\newblock The llama 3 herd of models, 2024.

\bibitem{he-etal-2024-softdedup}
Nan He, Weichen Xiong, Hanwen Liu, Yi~Liao, Lei Ding, Kai Zhang, Guohua Tang, Xiao Han, and Yang Wei.
\newblock {S}oft{D}edup: an efficient data reweighting method for speeding up language model pre-training.
\newblock In Lun-Wei Ku, Andre Martins, and Vivek Srikumar, editors, {\em Proceedings of the 62nd Annual Meeting of the Association for Computational Linguistics (Volume 1: Long Papers)}, pages 4011--4022, Bangkok, Thailand, August 2024. Association for Computational Linguistics.

\bibitem{he2024scalinglawsmultilinguallanguage}
Yifei He, Alon Benhaim, Barun Patra, Praneetha Vaddamanu, Sanchit Ahuja, Parul Chopra, Vishrav Chaudhary, Han Zhao, and Xia Song.
\newblock Scaling laws for multilingual language models, 2024.

\bibitem{henighan2020scalinglawsautoregressivegenerative}
Tom Henighan, Jared Kaplan, Mor Katz, Mark Chen, Christopher Hesse, Jacob Jackson, Heewoo Jun, Tom~B. Brown, Prafulla Dhariwal, Scott Gray, Chris Hallacy, Benjamin Mann, Alec Radford, Aditya Ramesh, Nick Ryder, Daniel~M. Ziegler, John Schulman, Dario Amodei, and Sam McCandlish.
\newblock Scaling laws for autoregressive generative modeling, 2020.

\bibitem{10.5555/3600270.3602446}
Jordan Hoffmann, Sebastian Borgeaud, Arthur Mensch, Elena Buchatskaya, Trevor Cai, Eliza Rutherford, Diego de~Las~Casas, Lisa~Anne Hendricks, Johannes Welbl, Aidan Clark, Tom Hennigan, Eric Noland, Katie Millican, George van~den Driessche, Bogdan Damoc, Aurelia Guy, Simon Osindero, Karen Simonyan, Erich Elsen, Oriol Vinyals, Jack~W. Rae, and Laurent Sifre.
\newblock Training compute-optimal large language models.
\newblock In {\em Proceedings of the 36th International Conference on Neural Information Processing Systems}, NIPS '22, Red Hook, NY, USA, 2022. Curran Associates Inc.

\bibitem{kang2025autoscalescaleawaredatamixing}
Feiyang Kang, Yifan Sun, Bingbing Wen, Si~Chen, Dawn Song, Rafid Mahmood, and Ruoxi Jia.
\newblock Autoscale: Scale-aware data mixing for pre-training llms, 2025.

\bibitem{kaplan2020scalinglawsneurallanguage}
Jared Kaplan, Sam McCandlish, Tom Henighan, Tom~B. Brown, Benjamin Chess, Rewon Child, Scott Gray, Alec Radford, Jeffrey Wu, and Dario Amodei.
\newblock Scaling laws for neural language models, 2020.

\bibitem{lai-etal-2024-llms}
Wen Lai, Mohsen Mesgar, and Alexander Fraser.
\newblock {LLM}s beyond {E}nglish: Scaling the multilingual capability of {LLM}s with cross-lingual feedback.
\newblock In Lun-Wei Ku, Andre Martins, and Vivek Srikumar, editors, {\em Findings of the Association for Computational Linguistics: ACL 2024}, pages 8186--8213, Bangkok, Thailand, August 2024. Association for Computational Linguistics.

\bibitem{li-etal-2024-cmmlu}
Haonan Li, Yixuan Zhang, Fajri Koto, Yifei Yang, Hai Zhao, Yeyun Gong, Nan Duan, and Timothy Baldwin.
\newblock {CMMLU}: Measuring massive multitask language understanding in {C}hinese.
\newblock In Lun-Wei Ku, Andre Martins, and Vivek Srikumar, editors, {\em Findings of the Association for Computational Linguistics: ACL 2024}, pages 11260--11285, Bangkok, Thailand, August 2024. Association for Computational Linguistics.

\bibitem{lin-etal-2022-truthfulqa}
Stephanie Lin, Jacob Hilton, and Owain Evans.
\newblock {T}ruthful{QA}: Measuring how models mimic human falsehoods.
\newblock In Smaranda Muresan, Preslav Nakov, and Aline Villavicencio, editors, {\em Proceedings of the 60th Annual Meeting of the Association for Computational Linguistics (Volume 1: Long Papers)}, pages 3214--3252, Dublin, Ireland, May 2022. Association for Computational Linguistics.

\bibitem{lin-etal-2022-shot}
Xi~Victoria Lin, Todor Mihaylov, Mikel Artetxe, Tianlu Wang, Shuohui Chen, Daniel Simig, Myle Ott, Naman Goyal, Shruti Bhosale, Jingfei Du, Ramakanth Pasunuru, Sam Shleifer, Punit~Singh Koura, Vishrav Chaudhary, Brian O{'}Horo, Jeff Wang, Luke Zettlemoyer, Zornitsa Kozareva, Mona Diab, Veselin Stoyanov, and Xian Li.
\newblock Few-shot learning with multilingual generative language models.
\newblock In Yoav Goldberg, Zornitsa Kozareva, and Yue Zhang, editors, {\em Proceedings of the 2022 Conference on Empirical Methods in Natural Language Processing}, pages 9019--9052, Abu Dhabi, United Arab Emirates, December 2022. Association for Computational Linguistics.

\bibitem{NEURIPS2024_3322a9a7}
Zhenghao Lin, Zhibin Gou, Yeyun Gong, Xiao Liu, Yelong Shen, Ruochen Xu, Chen Lin, Yujiu Yang, Jian Jiao, Nan Duan, and Weizhu Chen.
\newblock Not all tokens are what you need for pretraining.
\newblock In A.~Globerson, L.~Mackey, D.~Belgrave, A.~Fan, U.~Paquet, J.~Tomczak, and C.~Zhang, editors, {\em Advances in Neural Information Processing Systems}, volume~37, pages 29029--29063. Curran Associates, Inc., 2024.

\bibitem{liu2025regmix}
Qian Liu, Xiaosen Zheng, Niklas Muennighoff, Guangtao Zeng, Longxu Dou, Tianyu Pang, Jing Jiang, and Min Lin.
\newblock Regmix: Data mixture as regression for language model pre-training.
\newblock In {\em The Thirteenth International Conference on Learning Representations}, 2025.

\bibitem{pmlr-v235-ludziejewski24a}
Jan Ludziejewski, Jakub Krajewski, Kamil Adamczewski, Maciej Pi\'{o}ro, Micha{\l} Krutul, Szymon Antoniak, Kamil Ciebiera, Krystian Kr\'{o}l, Tomasz Odrzyg\'{o}\'{z}d\'{z}, Piotr Sankowski, Marek Cygan, and Sebastian Jaszczur.
\newblock Scaling laws for fine-grained mixture of experts.
\newblock In Ruslan Salakhutdinov, Zico Kolter, Katherine Heller, Adrian Weller, Nuria Oliver, Jonathan Scarlett, and Felix Berkenkamp, editors, {\em Proceedings of the 41st International Conference on Machine Learning}, volume 235 of {\em Proceedings of Machine Learning Research}, pages 33270--33288. PMLR, 21--27 Jul 2024.

\bibitem{muennighoff-etal-2023-crosslingual}
Niklas Muennighoff, Thomas Wang, Lintang Sutawika, Adam Roberts, Stella Biderman, Teven Le~Scao, M~Saiful Bari, Sheng Shen, Zheng~Xin Yong, Hailey Schoelkopf, Xiangru Tang, Dragomir Radev, Alham~Fikri Aji, Khalid Almubarak, Samuel Albanie, Zaid Alyafeai, Albert Webson, Edward Raff, and Colin Raffel.
\newblock Crosslingual generalization through multitask finetuning.
\newblock In Anna Rogers, Jordan Boyd-Graber, and Naoaki Okazaki, editors, {\em Proceedings of the 61st Annual Meeting of the Association for Computational Linguistics (Volume 1: Long Papers)}, pages 15991--16111, Toronto, Canada, July 2023. Association for Computational Linguistics.

\bibitem{openai2024gpt4technicalreport}
OpenAI.
\newblock Gpt-4 technical report, 2024.

\bibitem{openai2024gpt4ocard}
OpenAI.
\newblock Gpt-4o system card, 2024.

\bibitem{penedo2024fineweb-2}
Guilherme Penedo, Hynek Kydlíček, Vinko Sabolčec, Bettina Messmer, Negar Foroutan, Martin Jaggi, Leandro von Werra, and Thomas Wolf.
\newblock Fineweb2: A sparkling update with 1000s of languages, December 2024.

\bibitem{ponti-etal-2020-xcopa}
Edoardo~Maria Ponti, Goran Glava{\v{s}}, Olga Majewska, Qianchu Liu, Ivan Vuli{\'c}, and Anna Korhonen.
\newblock {XCOPA}: A multilingual dataset for causal commonsense reasoning.
\newblock In Bonnie Webber, Trevor Cohn, Yulan He, and Yang Liu, editors, {\em Proceedings of the 2020 Conference on Empirical Methods in Natural Language Processing (EMNLP)}, pages 2362--2376, Online, November 2020. Association for Computational Linguistics.

\bibitem{qin2025scalinglawssyntheticdata}
Zeyu Qin, Qingxiu Dong, Xingxing Zhang, Li~Dong, Xiaolong Huang, Ziyi Yang, Mahmoud Khademi, Dongdong Zhang, Hany~Hassan Awadalla, Yi~R. Fung, Weizhu Chen, Minhao Cheng, and Furu Wei.
\newblock Scaling laws of synthetic data for language models, 2025.

\bibitem{bai2023qwentechnicalreport}
Qwen.
\newblock Qwen technical report, 2023.

\bibitem{qwen2025qwen25technicalreport}
Qwen.
\newblock Qwen2.5 technical report, 2025.

\bibitem{rein2024gpqa}
David Rein, Betty~Li Hou, Asa~Cooper Stickland, Jackson Petty, Richard~Yuanzhe Pang, Julien Dirani, Julian Michael, and Samuel~R. Bowman.
\newblock {GPQA}: A graduate-level google-proof q\&a benchmark.
\newblock In {\em First Conference on Language Modeling}, 2024.

\bibitem{singh2025globalmmluunderstandingaddressing}
Shivalika Singh, Angelika Romanou, Clémentine Fourrier, David~I. Adelani, Jian~Gang Ngui, Daniel Vila-Suero, Peerat Limkonchotiwat, Kelly Marchisio, Wei~Qi Leong, Yosephine Susanto, Raymond Ng, Shayne Longpre, Wei-Yin Ko, Sebastian Ruder, Madeline Smith, Antoine Bosselut, Alice Oh, Andre F.~T. Martins, Leshem Choshen, Daphne Ippolito, Enzo Ferrante, Marzieh Fadaee, Beyza Ermis, and Sara Hooker.
\newblock Global mmlu: Understanding and addressing cultural and linguistic biases in multilingual evaluation, 2025.

\bibitem{10.5555/3600270.3601689}
Ben Sorscher, Robert Geirhos, Shashank Shekhar, Surya Ganguli, and Ari~S. Morcos.
\newblock Beyond neural scaling laws: beating power law scaling via data pruning.
\newblock In {\em Proceedings of the 36th International Conference on Neural Information Processing Systems}, NIPS '22, Red Hook, NY, USA, 2022. Curran Associates Inc.

\bibitem{sow2025dynamic}
Daouda Sow, Herbert Woisetschl{\"a}ger, Saikiran Bulusu, Shiqiang Wang, Hans~Arno Jacobsen, and Yingbin Liang.
\newblock Dynamic loss-based sample reweighting for improved large language model pretraining.
\newblock In {\em The Thirteenth International Conference on Learning Representations}, 2025.

\bibitem{nllbteam2022languageleftbehindscaling}
NLLB Team, Marta~R. Costa-jussà, James Cross, Onur Çelebi, Maha Elbayad, Kenneth Heafield, Kevin Heffernan, Elahe Kalbassi, Janice Lam, Daniel Licht, Jean Maillard, Anna Sun, Skyler Wang, Guillaume Wenzek, Al~Youngblood, Bapi Akula, Loic Barrault, Gabriel~Mejia Gonzalez, Prangthip Hansanti, John Hoffman, Semarley Jarrett, Kaushik~Ram Sadagopan, Dirk Rowe, Shannon Spruit, Chau Tran, Pierre Andrews, Necip~Fazil Ayan, Shruti Bhosale, Sergey Edunov, Angela Fan, Cynthia Gao, Vedanuj Goswami, Francisco Guzmán, Philipp Koehn, Alexandre Mourachko, Christophe Ropers, Safiyyah Saleem, Holger Schwenk, and Jeff Wang.
\newblock No language left behind: Scaling human-centered machine translation, 2022.

\bibitem{yang2025qwen3technicalreport}
Qwen Team.
\newblock Qwen3 technical report, 2025.

\bibitem{thakkar-etal-2023-self}
Megh Thakkar, Tolga Bolukbasi, Sriram Ganapathy, Shikhar Vashishth, Sarath Chandar, and Partha Talukdar.
\newblock Self-influence guided data reweighting for language model pre-training.
\newblock In Houda Bouamor, Juan Pino, and Kalika Bali, editors, {\em Proceedings of the 2023 Conference on Empirical Methods in Natural Language Processing}, pages 2033--2045, Singapore, December 2023. Association for Computational Linguistics.

\bibitem{touvron2023llama2openfoundation}
Hugo Touvron and Louis~Martin et~al.
\newblock Llama 2: Open foundation and fine-tuned chat models, 2023.

\bibitem{touvron2023llamaopenefficientfoundation}
Hugo Touvron, Thibaut Lavril, Gautier Izacard, Xavier Martinet, Marie-Anne Lachaux, Timothée Lacroix, Baptiste Rozière, Naman Goyal, Eric Hambro, Faisal Azhar, Aurelien Rodriguez, Armand Joulin, Edouard Grave, and Guillaume Lample.
\newblock Llama: Open and efficient foundation language models, 2023.

\bibitem{10890139}
Dominik Wagner, Alexander Churchill, Siddharth Sigtia, and Erik Marchi.
\newblock Selma: A speech-enabled language model for virtual assistant interactions.
\newblock In {\em ICASSP 2025 - 2025 IEEE International Conference on Acoustics, Speech and Signal Processing (ICASSP)}, pages 1--5, 2025.

\bibitem{NEURIPS2024_ed165f2f}
Jiachen~T. Wang, Tong Wu, Dawn Song, Prateek Mittal, and Ruoxi Jia.
\newblock Greats: Online selection of high-quality data for llm training in every iteration.
\newblock In A.~Globerson, L.~Mackey, D.~Belgrave, A.~Fan, U.~Paquet, J.~Tomczak, and C.~Zhang, editors, {\em Advances in Neural Information Processing Systems}, volume~37, pages 131197--131223. Curran Associates, Inc., 2024.

\bibitem{NEURIPS2023_dcba6be9}
Sang~Michael Xie, Hieu Pham, Xuanyi Dong, Nan Du, Hanxiao Liu, Yifeng Lu, Percy~S Liang, Quoc~V Le, Tengyu Ma, and Adams~Wei Yu.
\newblock Doremi: Optimizing data mixtures speeds up language model pretraining.
\newblock In A.~Oh, T.~Naumann, A.~Globerson, K.~Saenko, M.~Hardt, and S.~Levine, editors, {\em Advances in Neural Information Processing Systems}, volume~36, pages 69798--69818. Curran Associates, Inc., 2023.

\bibitem{ye2025data}
Jiasheng Ye, Peiju Liu, Tianxiang Sun, Jun Zhan, Yunhua Zhou, and Xipeng Qiu.
\newblock Data mixing laws: Optimizing data mixtures by predicting language modeling performance.
\newblock In {\em The Thirteenth International Conference on Learning Representations}, 2025.

\bibitem{NEURIPS2024_c4bec0d2}
Zichun Yu, Spandan Das, and Chenyan Xiong.
\newblock Mates: Model-aware data selection for efficient pretraining with data influence models.
\newblock In A.~Globerson, L.~Mackey, D.~Belgrave, A.~Fan, U.~Paquet, J.~Tomczak, and C.~Zhang, editors, {\em Advances in Neural Information Processing Systems}, volume~37, pages 108735--108759. Curran Associates, Inc., 2024.

\bibitem{zellers-etal-2019-hellaswag}
Rowan Zellers, Ari Holtzman, Yonatan Bisk, Ali Farhadi, and Yejin Choi.
\newblock {H}ella{S}wag: Can a machine really finish your sentence?
\newblock In Anna Korhonen, David Traum, and Llu{\'i}s M{\`a}rquez, editors, {\em Proceedings of the 57th Annual Meeting of the Association for Computational Linguistics}, pages 4791--4800, Florence, Italy, July 2019. Association for Computational Linguistics.

\bibitem{zhang2025harnessing}
Chi Zhang, Huaping Zhong, Kuan Zhang, Chengliang Chai, Rui Wang, Xinlin Zhuang, Tianyi Bai, Qiu Jiantao, Lei Cao, Ju~Fan, Ye~Yuan, Guoren Wang, and Conghui He.
\newblock Harnessing diversity for important data selection in pretraining large language models.
\newblock In {\em The Thirteenth International Conference on Learning Representations}, 2025.

\bibitem{zhu2024multilinguallargelanguagemodels}
Shaolin Zhu, Supryadi, Shaoyang Xu, Haoran Sun, Leiyu Pan, Menglong Cui, Jiangcun Du, Renren Jin, António Branco, and Deyi Xiong.
\newblock Multilingual large language models: A systematic survey, 2024.

\bibitem{zhu-etal-2024-multilingual}
Wenhao Zhu, Hongyi Liu, Qingxiu Dong, Jingjing Xu, Shujian Huang, Lingpeng Kong, Jiajun Chen, and Lei Li.
\newblock Multilingual machine translation with large language models: Empirical results and analysis.
\newblock In Kevin Duh, Helena Gomez, and Steven Bethard, editors, {\em Findings of the Association for Computational Linguistics: NAACL 2024}, pages 2765--2781, Mexico City, Mexico, June 2024. Association for Computational Linguistics.

\bibitem{zhu2025toremitopicawaredatareweighting}
Xiaoxuan Zhu, Zhouhong Gu, Baiqian Wu, Suhang Zheng, Tao Wang, Tianyu Li, Hongwei Feng, and Yanghua Xiao.
\newblock Toremi: Topic-aware data reweighting for dynamic pre-training data selection, 2025.

\bibitem{zhuocheng-etal-2023-scaling}
Zhang Zhuocheng, Shuhao Gu, Min Zhang, and Yang Feng.
\newblock Scaling law for document neural machine translation.
\newblock In Houda Bouamor, Juan Pino, and Kalika Bali, editors, {\em Findings of the Association for Computational Linguistics: EMNLP 2023}, pages 8290--8303, Singapore, December 2023. Association for Computational Linguistics.

\end{thebibliography}

\newpage
\appendix

\section{Details of Fitting Transfer Strength \(\alpha_{j\rightarrow i}(D)\)}
\label{appendix:alpha_fitting}
\setcounter{figure}{8}
\begin{figure}[ht]
    \centering
    \subfigure[$\alpha_{ar\rightarrow en}$ as a function of $D$]{
        \includegraphics[width=0.3\textwidth]{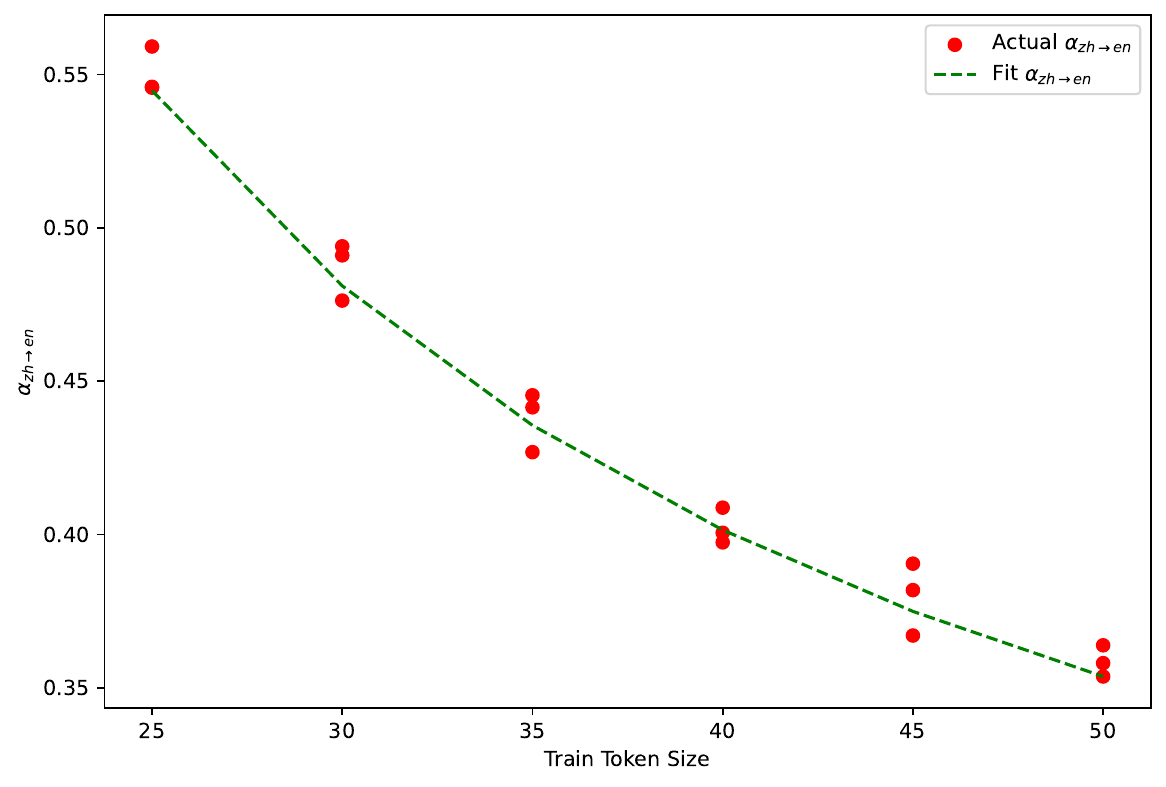}
    }\hfill
    \subfigure[$\alpha_{ar\rightarrow ko}$ as a function of $D$]{
        \includegraphics[width=0.3\textwidth]{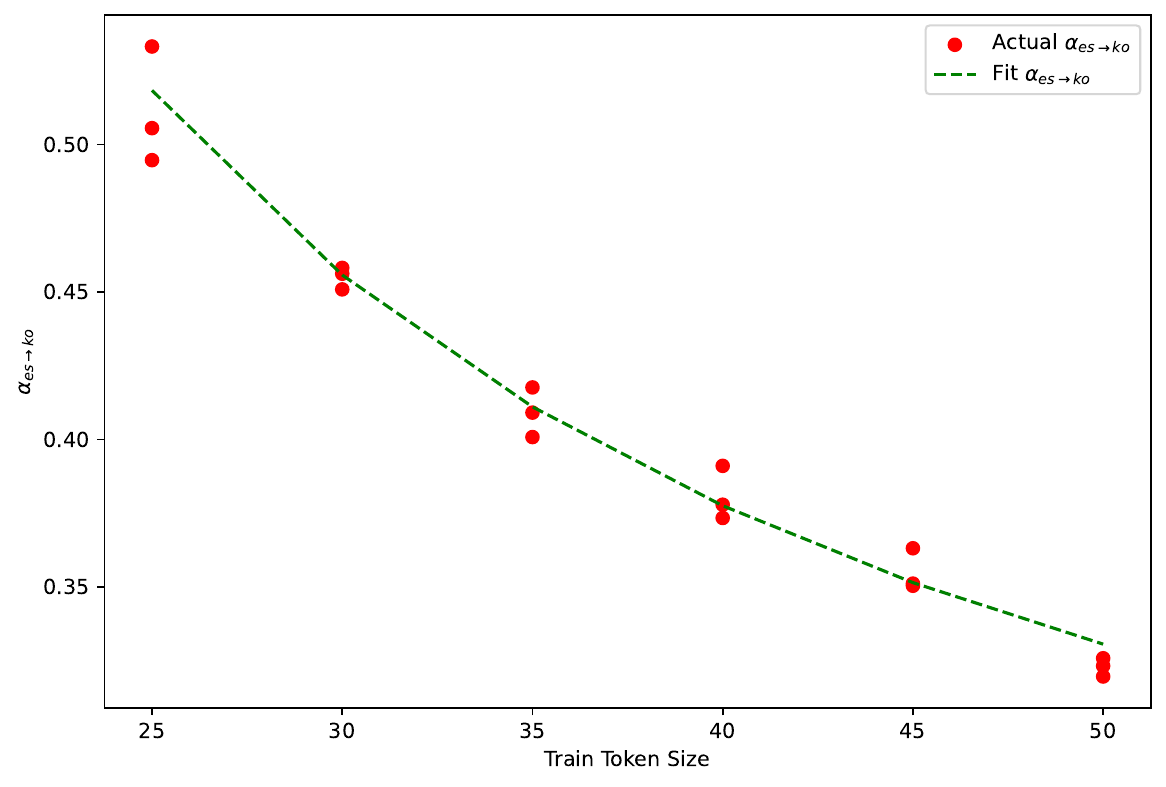}
    }\hfill
    \subfigure[$\alpha_{ar\rightarrow zh}$ as a function of $D$]{
        \includegraphics[width=0.3\textwidth]{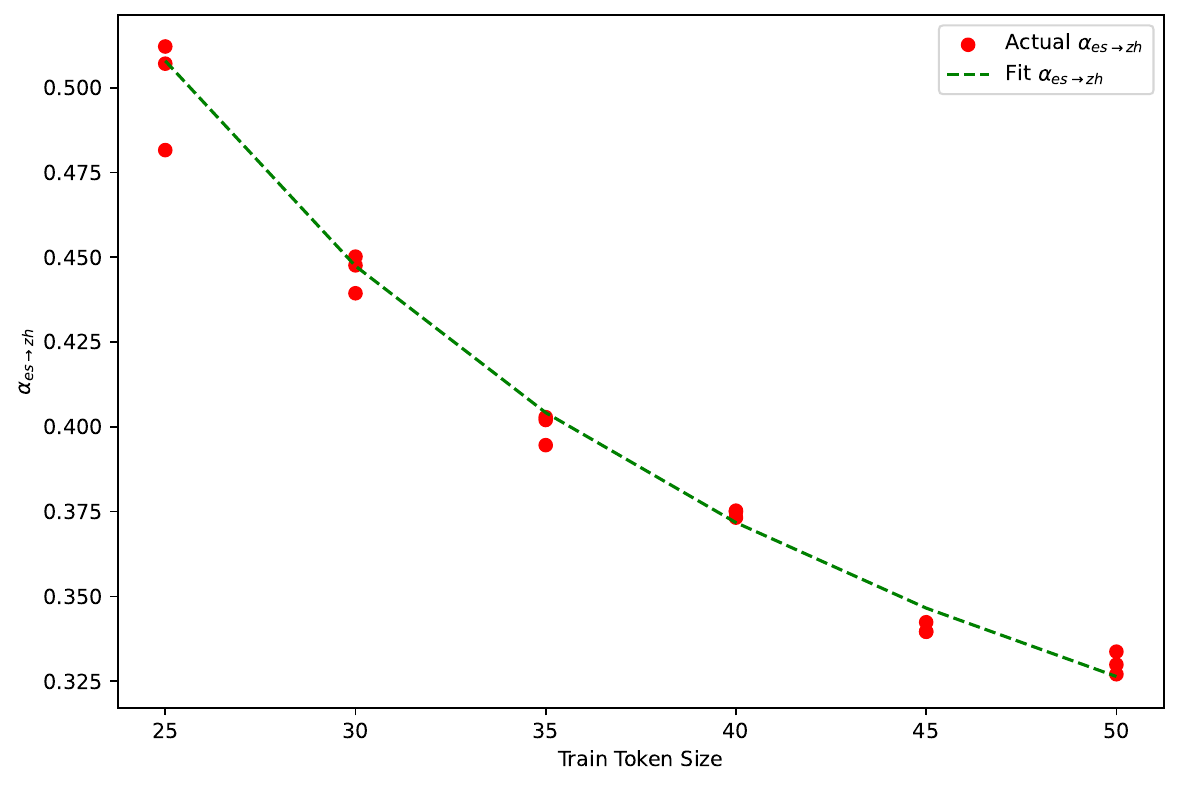}
    }
    \caption{Illustration of cross-lingual interaction-aware language ratio ($\tilde{r}_{ar}$) and its dependency on original training proportions ($r_{ar}$).}
    \label{fig:alpha_vs_D}
\end{figure}

As introduced in Figure~2 (c) of the main text, we observe that the curve relating \(\tilde{r}_i\) and \(r_i\) shifts vertically depending on the total token budget \(D\). Specifically, as \(D\) increases, the \(\tilde{r}_i\) versus \(r_i\) curve tends to move downward, while smaller \(D\) values correspond to upward shifts. According to Equation (4), the parameter \(\alpha_{j\rightarrow i}(D)\) effectively acts as an intercept controlling this vertical shift.

To accurately characterize the relationship between \(\alpha_{j\rightarrow i}\) and the data budget \(D\), we adopt a two-step procedure. First, we individually fit the relationship between \(\tilde{r}_i\) and \(r_i\) at different values of \(D\) using Equation (4). This yields empirical estimates of \(\alpha_{j\rightarrow i}\) at various token budgets. Figure~\ref{fig:alpha_vs_D} illustrates the computed values of \(\alpha_{j\rightarrow i}\) for three representative language pairs across different scales of \(D\).

Moreover, our empirical findings suggest two critical properties for the \(\alpha_{j\rightarrow i}(D)\) relationship:
\begin{itemize}
    \item \textbf{Non-monotonicity}: \(\alpha_{j\rightarrow i}\) does not continuously decrease with increasing \(D\); rather, it converges towards a stable limiting value as \(D\) becomes sufficiently large.
    \item \textbf{Sign variability}: \(\alpha_{j\rightarrow i}\) can be either positive or negative. Positive values indicate beneficial cross-lingual transfer, whereas negative values reflect interference effects, where additional data from language \(L_j\) eventually hinder the learning of language \(L_i\).
\end{itemize}

Considering these empirical insights, we propose modeling \(\alpha_{j\rightarrow i}(D)\) with the following parametric form:
\begin{equation}
    \alpha_{j\rightarrow i}(D) = b_{ji} + \frac{k_{ji}}{D},
\end{equation}
where \(b_{ji}\) represents the asymptotic transfer strength as \(D\to\infty\), and \(k_{ji}\) controls the decay rate of this transfer effect as the data budget increases.

The fitting results using this parametric form, depicted by the green curves in Figure~\ref{fig:alpha_vs_D}, demonstrate excellent agreement with the empirical \(\alpha_{j\rightarrow i}\)-\(D\) relationships across various language pairs, validating our choice of functional form.

\section{Derivation of Optimal Direction for Cross-Lingual Interaction-Aware Ratios \(p_i\)}
\label{appendix:optimal_direction_derivation}

To compute the optimal direction of the Cross-Lingual Interaction-Aware Ratios \(\{\tilde r_i\}\), we formulate and solve the following uncoupled optimization subproblem:
\begin{equation}
\min_{\tilde r_i>0}\;\sum_{i=1}^n \frac{B_i}{(D\,\tilde r_i)^{\beta_i}}
\quad\text{s.t.}\quad
\sum_{i=1}^n \tilde r_i = M,
\end{equation}
where \(M>0\) is a fixed normalization constant, and \(B_i,\beta_i,D\) are known positive parameters.

Introducing a Lagrange multiplier \(\lambda\), we construct the Lagrangian:
\begin{equation}
\mathcal{J}(\tilde{\mathbf r},\lambda)
=
\sum_{i=1}^n \frac{B_i}{(D\,\tilde r_i)^{\beta_i}}
\;+\;\lambda\left(\sum_{i=1}^n\tilde r_i - M\right).
\end{equation}
Taking derivatives with respect to each \(\tilde r_i\) and setting them to zero, we obtain the first-order optimality conditions:
\begin{align}
-\,B_i\,\beta_i\,D^{-\beta_i}\,\tilde r_i^{-(\beta_i+1)} + \lambda &= 0 \\
\Longrightarrow \quad
\tilde r_i^{\,\beta_i+1}
&= \frac{B_i\,\beta_i}{\lambda\,D^{\beta_i}}.
\end{align}

Comparing the conditions for any two languages \(i,j\), we have:
\begin{equation}
\frac{\tilde r_i}{\tilde r_j}
= \left(\frac{B_i\,\beta_i}{B_j\,\beta_j}\,D^{\beta_j-\beta_i}\right)^{\frac{1}{\beta_i+1}\big/\frac{1}{\beta_j+1}}.
\end{equation}
Thus, the optimal direction must satisfy:
\begin{equation}
\tilde r_i \;\propto\;
\left(B_i\,\beta_i / D^{\beta_i}\right)^{1/(\beta_i+1)}.
\end{equation}

Applying the normalization constraint \(\sum_i\tilde r_i=M\), we obtain the normalized optimal direction:
\begin{equation}
p_i = \frac{(B_i\,\beta_i)^{1/(\beta_i+1)}\,D^{-\beta_i/(\beta_i+1)}}
{\sum_{k=1}^n (B_k\,\beta_k)^{1/(\beta_k+1)}\,D^{-\beta_k/(\beta_k+1)}}.
\end{equation}

Since each term \(B_i/(D\tilde r_i)^{\beta_i}\) is strictly convex in \(\tilde r_i\) and the constraint is linear, the stationary solution derived above constitutes the unique global minimizer. This rigorous derivation justifies the Marginal-Benefit Balancing approach presented in the main text, providing the closed-form solution for the optimal direction \(\{\tilde r_i\}\).

\section{Equivalence of Two-Stage Optimization with Direct Optimization}
\label{appendix:optimization_equivalence}

Here we provide a rigorous justification demonstrating that our proposed two-stage optimization approach—first determining the optimal direction \(p_i\) and subsequently maximizing the magnitude of effective data allocation—is equivalent to directly solving the original optimization problem.

\textbf{(i) Necessity of Optimizing the Direction:} Assume the direction of the cross-lingual interaction-aware ratios \(\{\tilde{r}_i\}\) deviates from the optimal direction \(p_i\). Under any fixed effective data contribution \(\sum_i B_i / (D \tilde{r}_i)^{\beta_i}\), the total validation loss will always be greater than or equal to that obtained using the optimal direction. Formally, the optimal direction condition is:
\begin{equation}
\frac{B_i \beta_i}{D^{\beta_i} \tilde{r}_i^{\beta_i+1}} = \frac{B_j \beta_j}{D^{\beta_j} \tilde{r}_j^{\beta_j+1}}, \quad \forall i,j.
\end{equation}
Any deviation from this balanced proportionality condition disrupts marginal equilibrium, causing certain languages to have unnecessarily higher marginal loss reductions, thus reducing overall efficiency. Hence, identifying the direction \(\{\tilde{r}_i\}\) by balancing marginal benefits ensures minimal total loss given a fixed effective data contribution.

\textbf{(ii) Optimal Magnitude via Maximizing Effective Allocation:} Once the optimal direction \(p_i\) is fixed, we set \(\tilde{r}_i = c \cdot p_i\), where \(c\) denotes the scaling magnitude of effective data allocation (with normalization \(\sum_i \tilde{r}_i = c\)). We then isolate the variable component of total loss as a function of \(c\):
\begin{equation}
L_{\text{var}}(c) = \sum_i \frac{B_i}{(D c p_i)^{\beta_i}} = \sum_i \frac{B_i}{D^{\beta_i} p_i^{\beta_i}} c^{-\beta_i}.
\end{equation}

Differentiating with respect to \(c\), we have:
\begin{equation}
\frac{d L_{\text{var}}}{dc} = -\sum_i \frac{\beta_i B_i}{D^{\beta_i} p_i^{\beta_i}} c^{-(\beta_i+1)} < 0,
\end{equation}
provided that all \(\beta_i > 0\). This negative derivative demonstrates a strictly monotonic decrease in loss as the magnitude \(c\) increases. Intuitively, larger \(c\) means greater effective data volumes \(D \tilde{r}_i\) for each language, which consistently reduces loss due to the monotonicity of scaling laws. Therefore, to minimize the loss, we naturally aim to increase \(c\) as much as feasible—maximizing the total effective data contribution while maintaining the optimal relative proportions.

However, practical constraints limit the maximum achievable \(c\). Given the normalization constraint \(\sum r_i = 1\) and the implicit mapping from \(\{r_i\}\) to \(\{\tilde{r}_i\}\), the magnitude \(c\) has an upper bound \(c^*\) corresponding to feasible allocations.

In summary, stage 1 guarantees that adjusting the direction of ratios does not increase the loss, and stage 2 optimally maximizes effective data volume along this direction, ensuring minimal achievable loss. Thus, the two-stage solution is proven equivalent to directly solving the original optimization problem. This result aligns with previous studies on multilingual scaling laws, demonstrating the consistency and optimality of the two-stage optimization procedure.

\section{Training Details}
\label{appendix:training_details}

\textbf{Dataset Description}

All experiments utilize data sampled from the Fineweb-2 corpus \cite{penedo2024fineweb-2}. We further preprocess the dataset by training a custom Byte-Pair Encoding (BPE) tokenizer using the BBPE method, resulting in a vocabulary of 250k tokens for subsequent experiments.

\textbf{Experimental Setup}

We conduct multilingual experiments with various language combinations:
\begin{itemize}
    \item \textbf{Bilingual Experiments:} \{es-ko, en-zh, de-ar, ko-ja\}
    \item \textbf{Trilingual Experiments:} \{es-de-ar, es-ko-zh, en-zh-ja\}
    \item \textbf{Five-language Experiment:} \{es-de-ar-ko-ja\}
    \item \textbf{Sixteen-language Experiment:} \{de, en, nl, es, pt, fr, it, id, ja, ko, zh, ru, ar, th, vi, tr\}
\end{itemize}

As detailed in Algorithm 1, for each multilingual setting, we first fix the proportion of one language and evenly distribute the remaining proportion among the other languages. For each selected language \(L_i\), we systematically vary its proportion across the set \{0.02, 0.025, 0.05, 0.1, 0.2, 0.25, 0.4, 0.5, 0.6, 0.75, 0.8, 0.9, 0.95, 0.975, 0.98\} to establish comprehensive fitting functions. In the sixteen-language experiment, we follow Algorithm 1 for extrapolation and validation.

\textbf{Model Configuration}

We adopt a transformer-based architecture inspired by the LLaMA-2 \citep{touvron2023llama2openfoundation} model, specifically configured with approximately 1.2 billion parameters. The detailed architecture settings are:
\begin{itemize}
    \item Hidden size: 2048
    \item Vocabulary embedding dimension: 2048
    \item Intermediate layer dimension: 5504
    \item Attention heads: 16
    \item Layers: 24
    \item Maximum positional embeddings: 4096
    \item Layer normalization epsilon: \(1.0 \times 10^{-5}\)
\end{itemize}

All models are randomly initialized.

\textbf{Training Hyperparameters}

\begin{itemize}
    \item Batch size: 3072
    \item Sequence length: 4096
    \item Optimizer: AdamW
    \item Learning rate schedule: Cosine decay to 10\% of initial value
    \item Training steps: Varied according to total token budget \(D\)
    \item Precision: bf16 (mixed-precision training)
\end{itemize}

\textbf{Computational Resources and Runtime}

Each experiment is conducted using 64 H100 GPUs, with an average runtime of approximately 10 hours per experiment.

\textbf{Evaluation Methodology}

The validation datasets for each language are separately sampled from Fineweb-2, ensuring no overlap with training samples. Validation loss is computed by averaging the loss across the final three training steps of each run.

\section{Detailed Evaluation Protocols for Benchmarks}
\label{appendix:evaluation_protocols}

To rigorously assess the capabilities of our proposed model, we select benchmarks that span diverse evaluation dimensions, including natural language inference, commonsense reasoning, question answering, multilingual multitask understanding, and translation tasks. Recognizing that several benchmarks were originally developed only in English, we manually translated these datasets into multilingual versions (marked as \textsuperscript{\ddag}: XHS\textsuperscript{\ddag}, XARC-E\textsuperscript{\ddag}, XARC-C\textsuperscript{\ddag}, XGPQA\textsuperscript{\ddag}, XTQA\textsuperscript{\ddag}). Our translation approach involves encapsulating each benchmark component—prompts, questions, and answer choices—with explicit tags to maintain structural consistency. We then use GPT-based translation, ensuring strict validation of tag integrity post-translation, with any problematic samples retranslated. This systematic methodology guarantees accurate and faithful translations, supporting flexible future adaptations and mixed-language test scenarios. Below, we detail each evaluation benchmark grouped by task type.

\textbf{Language Modeling and Natural Language Inference}


\textit{XNLI (Cross-lingual Natural Language Inference)}~\citep{conneau-etal-2018-xnli}: Extended from MultiNLI, XNLI evaluates cross-lingual sentence representations across 15 languages, measuring models' inference capabilities.

\textit{XCOPA (Cross-lingual Choice of Plausible Alternatives)}~\citep{ponti-etal-2020-xcopa}: XCOPA tests models on causal commonsense reasoning across 11 languages, providing insights into multilingual causal reasoning capabilities.

\textit{XStoryCloze}~\citep{lin-etal-2022-shot}: XStoryCloze assesses zero-shot and few-shot learning across 10 non-English languages, examining models' narrative understanding and inference skills.

\textbf{Commonsense Reasoning}

\textit{HellaSwag (XHS\textsuperscript{\ddag})}~\citep{zellers-etal-2019-hellaswag}: Originally English-only, HellaSwag involves selecting the most plausible sentence ending from multiple choices, thereby testing commonsense reasoning.

\textit{XWinograd}~\citep{muennighoff-etal-2023-crosslingual}: As a multilingual variant of the Winograd Schema Challenge, XWinograd evaluates pronoun resolution abilities in diverse linguistic contexts.

\textbf{Question Answering}

\textit{ARC-Easy (XARC-E\textsuperscript{\ddag}) / ARC-Challenge (XARC-C\textsuperscript{\ddag})}~\citep{clark2018thinksolvedquestionanswering}: ARC contains scientific multiple-choice questions designed for different complexity levels, evaluating reasoning from basic to advanced.

\textit{GPQA (Graduate-Level Google-Proof Q\&A, XGPQA\textsuperscript{\ddag})} \citep{rein2024gpqa}: GPQA tests graduate-level understanding across domains like biology, physics, and chemistry, requiring deep comprehension beyond search-engine-based answers.

\textit{TruthfulQA (XTQA\textsuperscript{\ddag})} \citep{lin-etal-2022-truthfulqa}: This dataset assesses the factual accuracy and common misconception avoidance of language models across diverse topics.

\textbf{Multitask Language Understanding (MMLU Series)}

\textit{CMMLU (Chinese Massive Multitask Language Understanding)} \citep{li-etal-2024-cmmlu}: Evaluates Chinese language models' knowledge across multiple disciplines including natural sciences, engineering, and humanities.

\textit{JMMLU (Japanese Massive Multitask Language Understanding)} \footnote{\url{https://github.com/nlp-waseda/JMMLU}}: JMMLU assesses Japanese models on multitask language understanding, covering extensive topics.

\textit{VMLU (Vietnamese Massive Language Understanding)} \footnote{\url{https://vmlu.ai/}}: Focused on Vietnamese, VMLU evaluates broad academic and practical knowledge via a large set of multiple-choice questions.

\textit{GMMLU (Global Massive Multitask Language Understanding)} \citep{singh2025globalmmluunderstandingaddressing}: GMMLU tests multilingual generalization capabilities across various languages and diverse tasks.

\textbf{Translation Tasks}

\textit{FLORES (Facebook Low Resource Languages Evaluation Suite)} \citep{nllbteam2022languageleftbehindscaling}: Supporting many-to-many translations, FLORES provides a high-quality benchmark suitable for assessing model performance on low-resource languages.

\section{Detailed Per-Language Benchmark Results}
\label{appendix:per_language_results}

This appendix presents detailed, per-language evaluation results corresponding to the benchmarks summarized in Table~2. The following tables comprehensively report the performance of our \textsc{Climb}-derived multilingual allocation strategy across each evaluated language, facilitating an in-depth analysis and comparison against baseline methods.

\setcounter{table}{2} 

\begin{table}[htbp]
\centering
\caption{Detailed per-language performance on the \textbf{XWinograd} benchmark (5-shot accuracy).}
\begin{tabular}{@{}lcccccc@{}}
\toprule
\textbf{Model / Method} & \textbf{EN} & \textbf{FR} & \textbf{JP} & \textbf{PT} & \textbf{RU} & \textbf{ZH} \\
\midrule
\multicolumn{7}{l}{\textbf{Open Source Multilingual LLMs}} \\
\midrule
LLaMA-3.2 & 93.65 & 71.25 & 67.17 & 72.09 & 73.75 & 77.13 \\
Qwen-3 & 92.54 & 76.61 & 78.49 & 77.12 & 69.51 & 80.69 \\
Gemma-3 & 77.60 & 65.56 & 62.95 & 62.86 & 64.29 & 68.38 \\
\midrule
\multicolumn{7}{l}{\textbf{Different Data Allocation Methods}} \\
\midrule
Uniform & 82.93 & 72.45 & 71.39 & 73.80 & 67.89 & 71.58 \\
Isolated & 79.79 & 77.71 & 71.44 & 68.59 & 65.58 & 70.74 \\
Natural & 82.90 & 76.28 & 71.19 & 74.02 & 68.71 & \textbf{76.77} \\
MSL & 82.14 & 73.84 & 69.90 & 71.56 & 67.04 & 74.06 \\
\textsc{Climb} & \textbf{90.57} & \textbf{78.14} & \textbf{74.27} & \textbf{74.98} & \textbf{73.66} & 73.25 \\
\bottomrule
\end{tabular}
\label{tab:xwinograd_detailed_results}
\end{table}


\begin{table}[htbp]
\centering
\caption{Detailed per-language performance on the \textbf{XStoryCloze} benchmark (0-shot accuracy).}
\resizebox{\textwidth}{!}{%
\begin{tabular}{@{}lccccccccccc@{}}
\toprule
\textbf{Model / Method} & \textbf{AR} & \textbf{EN} & \textbf{ES} & \textbf{EU} & \textbf{HI} & \textbf{ID} & \textbf{MY} & \textbf{RU} & \textbf{SW} & \textbf{TE} & \textbf{ZH} \\
\midrule
\multicolumn{12}{l}{\textbf{Open Source Multilingual LLMs}} \\
\midrule
LLaMA-3.2 & 52.99 & 73.18 & 63.20 & 51.77 & 57.81 & 60.26 & 50.74 & 61.94 & 52.12 & 56.29 & 59.57 \\
Qwen-3 & 56.96 & 74.71 & 65.52 & 53.32 & 58.07 & 62.47 & 53.32 & 63.26 & 51.65 & 60.33 & 66.00 \\
Gemma-3 & 51.94 & 62.49 & 57.01 & 52.74 & 54.69 & 54.63 & 50.73 & 55.35 & 51.87 & 56.61 & 55.18 \\
\midrule
\multicolumn{12}{l}{\textbf{Different Data Allocation Methods}} \\
\midrule
Uniform & 60.45 & 70.35 & \bf 66.44 & 53.01 & 50.31 & \bf 65.22 & 49.98 & 65.25 & 51.19 & 54.91 & 61.76 \\
Isolated & 59.87 & 71.34 & 64.97 & 52.27 & 50.82 & 63.92 & 50.25 & 65.87 & 51.06 & 54.67 & 61.09 \\
Natural & 59.19 & 67.96 & 62.06 & 51.26 & 52.19 & 61.62 & 49.82 & 61.33 & 50.19 & 54.31 & 61.24 \\
MSL & 60.19 & 69.16 & 63.30 & 51.42 & 52.59 & 62.49 & 49.30 & 62.21 & 50.13 & 54.51 & 62.04 \\
\textsc{Climb} & \bf 62.39 & \bf 73.09 & 66.22 & \bf 53.74 & \bf 55.59 & 64.49 & \bf 51.11 & \bf 66.20 & \bf 52.57 & \bf 58.10 & \bf 62.43 \\
\bottomrule
\end{tabular}}
\label{tab:xstorycloze_detailed_results}
\end{table}


\begin{table}[htbp]
\centering
\caption{Detailed per-language performance on the \textbf{XCOPA} benchmark (5-shot accuracy).}
\resizebox{\textwidth}{!}{%
\begin{tabular}{@{}lccccccccccc@{}}
\toprule
\textbf{Model / Method} & \textbf{ET} & \textbf{HT} & \textbf{ID} & \textbf{IT} & \textbf{QU} & \textbf{SW} & \textbf{TA} & \textbf{TH} & \textbf{TR} & \textbf{VI} & \textbf{ZH} \\
\midrule
\multicolumn{12}{l}{\textbf{Open Source Multilingual LLMs}} \\
\midrule
LLaMA-3.2 & 52.09 & 52.31 & 62.69 & 62.49 & 51.51 & 51.31 & 55.08 & 55.90 & 55.90 & 64.68 & 64.47 \\
Qwen-3 & 52.57 & 53.17 & 66.63 & 65.07 & 49.77 & 53.17 & 54.38 & 57.81 & 57.81 & 70.03 & 74.64 \\
Gemma-3 & 51.99 & 52.77 & 60.18 & 56.59 & 52.20 & 55.21 & 55.62 & 54.19 & 55.62 & 59.79 & 57.99 \\
\midrule
\multicolumn{12}{l}{\textbf{Different Data Allocation Methods}} \\
\midrule
Uniform & 49.59 & 50.99 & 67.99 & \bf 66.99 & 51.63 & 51.63 & 56.46 & 61.05 & 61.26 & 69.59 & \bf 67.20 \\
Isolated & 49.86 & 51.66 & \bf 70.62 & 64.80 & 50.60 & 50.21 & 56.60 & \bf 61.78 & \bf 61.78 & \bf 69.94 & 65.77 \\
Natural & 50.76 & 51.48 & 64.31 & 59.09 & 49.97 & 51.85 & 54.44 & 58.76 & 58.49 & 62.85 & 59.93 \\
MSL & 52.32 & 52.77 & 66.29 & 61.15 & 51.18 & 52.88 & 55.45 & 59.54 & 60.00 & 64.75 & 62.33 \\
\textsc{Climb} & \bf 54.21 & \bf 53.80 & 68.06 & 63.89 & \bf 52.18 & \bf 54.12 & \bf 56.73 & 60.95 & 61.50 & 67.46 & 66.90 \\
\bottomrule
\end{tabular}}
\label{tab:xcopa_detailed_results}
\end{table}


\begin{table}[htbp]
\centering
\caption{Detailed per-language performance on the \textbf{XNLI} benchmark (5-shot accuracy).}
\resizebox{\textwidth}{!}{%
\begin{tabular}{@{}lcccccccccc@{}}
\toprule
\textbf{Model / Method} & \textbf{AR} & \textbf{DE} & \textbf{EN} & \textbf{ES} & \textbf{FR} & \textbf{RU} & \textbf{TH} & \textbf{TR} & \textbf{VI} & \textbf{ZH} \\
\midrule
\multicolumn{11}{l}{\textbf{Open Source Multilingual LLMs}} \\
\midrule
LLaMA-3.2 & 34.05 & 42.16 & 46.15 & 40.41 & 42.20 & 40.48 & 38.41 & 39.90 & 39.90 & 39.86 \\
Qwen-3 & 33.83 & 42.38 & 47.43 & 43.58 & 43.58 & 42.38 & 39.70 & 37.44 & 41.10 & 41.90 \\
Gemma-3 & 38.94 & 41.35 & 44.81 & 41.53 & 41.92 & 41.92 & 39.74 & 40.18 & 42.28 & 41.03 \\
\midrule
\multicolumn{11}{l}{\textbf{Different Data Allocation Methods}} \\
\midrule
Uniform & 32.68 & \bf 43.75 & 44.37 & 41.39 & 43.95 & 40.41 & 37.67 & \bf 41.81 & 36.45 & \bf 38.31 \\
Isolated & 31.15 & 40.95 & 42.76 & 40.16 & 42.84 & 40.22 & 36.53 & 41.60 & 35.64 & 37.46 \\
Natural & 34.26 & 40.61 & 43.19 & 40.23 & 41.53 & 39.40 & 37.50 & 39.58 & 35.74 & 38.46 \\
MSL & 32.88 & 39.57 & 43.00 & 40.23 & 40.95 & 38.88 & 37.62 & 38.66 & 35.47 & 38.14 \\
\textsc{Climb} & \bf 35.14 & 43.01 & \bf 48.18 & \bf 43.93 & \bf 44.41 & \bf 42.72 & \bf 40.87 & 41.76 & \bf 38.92 & 37.56 \\
\bottomrule
\end{tabular}}
\label{tab:xnli_detailed_results}
\end{table}


\begin{table}[htbp]
\centering
\caption{Detailed per-language performance on the \textbf{Global MMLU (GMMLU)} benchmark (5-shot accuracy).}
\resizebox{\textwidth}{!}{%
\begin{tabular}{@{}lcccccccccccccccc@{}}
\toprule
\textbf{Model / Method} & \textbf{AR} & \textbf{DE} & \textbf{EN} & \textbf{ES} & \textbf{FIL} & \textbf{FR} & \textbf{ID} & \textbf{IT} & \textbf{JA} & \textbf{KO} & \textbf{MS} & \textbf{NL} & \textbf{PT} & \textbf{TR} & \textbf{VI} & \textbf{ZH} \\
\midrule
\multicolumn{17}{l}{\textbf{Open Source Multilingual LLMs}} \\
\midrule
LLaMA-3.2 & 25.88 & 29.12 & 35.30 & 29.31 & 28.05 & 28.84 & 28.59 & 28.54 & 27.58 & 27.90 & 28.33 & 28.11 & 29.16 & 27.21 & 28.39 & 29.21 \\
Qwen-3 & 29.62 & 34.79 & 43.92 & 35.77 & 31.23 & 35.68 & 33.94 & 34.85 & 32.75 & 32.21 & 32.15 & 33.23 & 35.74 & 31.04 & 33.63 & 37.94 \\
Gemma-3 & 25.43 & 26.94 & 31.13 & 27.75 & 27.00 & 27.20 & 27.15 & 27.05 & 26.49 & 26.95 & 26.57 & 25.96 & 27.49 & 26.68 & 27.29 & 27.42 \\
\midrule
\multicolumn{17}{l}{\textbf{Different Data Allocation Methods}} \\
\midrule
Uniform & 27.56 & 29.78 & 31.30 & 29.81 & 25.64 & 29.85 & 29.76 & 29.37 & 28.45 & 28.77 & 28.51 & 28.88 & 30.18 & 28.75 & 28.99 & 29.20 \\
Isolated & 26.80 & 29.20 & 30.96 & 29.56 & 25.66 & 29.17 & 29.44 & 29.03 & 28.27 & 28.35 & 28.21 & 28.73 & 29.61 & 28.21 & 28.60 & 28.45 \\
Natural & 28.84 & 31.18 & 33.38 & 31.83 & 27.15 & 31.09 & 31.11 & 30.51 & 29.43 & 28.64 & 28.31 & 30.25 & 31.47 & 29.81 & 29.72 & 30.96 \\
MSL & 28.00 & 29.93 & 32.47 & 30.55 & 26.46 & 30.43 & 29.80 & 28.84 & 27.51 & 27.38 & 26.69 & 29.01 & 30.47 & 28.30 & 28.58 & \bf 29.60 \\
\textsc{Climb} & \bf 30.53 & \bf 32.91 & \bf 36.26 & \bf 33.85 & \bf 28.86 & \bf 33.95 & \bf 33.04 & \bf 32.11 & \bf 30.70 & \bf 30.33 & \bf 29.68 & \bf 31.48 & \bf 33.92 & \bf 30.79 & \bf 31.16 & 28.91 \\
\bottomrule
\end{tabular}}
\label{tab:global_mmlu_detailed_results}
\end{table}


\begin{table}[htbp]
\centering
\caption{Detailed per-language performance on the \textbf{FLORES Translation} benchmark (5-shot chrF++ scores).}
\resizebox{\textwidth}{!}{%
\begin{tabular}{@{}lcccccccccccccccccc@{}}
\toprule
\textbf{Model / Method} & \multicolumn{17}{c}{\textbf{Translation to English (xx-en)}} \\
\cmidrule(lr){2-18}
 & AR & DE & ES & FR & ID & IT & JA & KO & MS & NL & PT & RU & TH & TL & TR & VI & ZH \\
\midrule
LLaMA-3.2 & 46.47 & 57.15 & 51.99 & 58.86 & 53.57 & 53.20 & 39.51 & 39.20 & 52.00 & 50.82 & 61.54 & 50.82 & 42.53 & 42.56 & 42.26 & 48.77 & 44.12 \\
Qwen-3 & 55.40 & 61.66 & 54.54 & 62.48 & 58.90 & 56.20 & 48.68 & 48.82 & 58.09 & 53.48 & 64.18 & 55.65 & 49.75 & 51.95 & 51.39 & 54.26 & 52.12 \\
Gemma-3 & 43.14 & 53.40 & 38.32 & 49.88 & 40.80 & 46.72 & 36.66 & 28.78 & 42.00 & 43.14 & 52.78 & 45.00 & 35.26 & 37.90 & 38.32 & 38.06 & 40.29 \\
Uniform & 55.37 & 61.04 & 54.87 & 61.75 & 59.21 & 56.23 & 45.28 & 46.07 & 57.67 & 54.38 & 65.10 & 54.46 & 48.91 & 20.60 & 51.28 & 53.56 & 46.88 \\
Isolated & 55.57 & 60.91 & 54.24 & 61.77 & 59.62 & 55.73 & 45.86 & 46.43 & 57.82 & 54.92 & 64.74 & 54.64 & 49.19 & 20.68 & 51.81 & 53.53 & 46.38 \\
Natural & 56.99 & 61.56 & 55.39 & 62.87 & 60.74 & 56.99 & 46.47 & 47.01 & 58.48 & 54.79 & 65.69 & 55.60 & 49.82 & 22.59 & 52.61 & 54.99 & 47.57 \\
MSL & 56.15 & 60.65 & 54.35 & 61.85 & 59.94 & 56.15 & 45.96 & 46.43 & 57.34 & 53.97 & 64.40 & 54.47 & 48.76 & 23.12 & 51.95 & 54.23 & 47.02 \\
\textsc{Climb} & 58.99 & 63.65 & 57.13 & 65.55 & 63.43 & 59.08 & 48.89 & 49.32 & 60.26 & 56.75 & 66.66 & 57.12 & 51.36 & 25.46 & 54.77 & 57.01 & 46.89 \\
\midrule
\textbf{Model / Method} & \multicolumn{17}{c}{\textbf{Translation from English (en-xx)}} \\
\cmidrule(lr){2-18}
 & AR & DE & ES & FR & ID & IT & JA & KO & MS & NL & PT & RU & TH & TL & TR & VI & ZH \\
\midrule
LLaMA-3.2 & 27.93 & 49.67 & 47.56 & 55.84 & 52.44 & 45.92 & 19.61 & 17.24 & 47.10 & 46.11 & 57.57 & 42.05 & 25.85 & 30.45 & 33.82 & 45.69 & 20.53 \\
Qwen-3 & 36.82 & 54.06 & 50.86 & 61.53 & 59.39 & 50.22 & 26.69 & 23.64 & 52.19 & 46.81 & 62.44 & 47.53 & 35.09 & 37.32 & 39.47 & 53.24 & 30.23 \\
Gemma-3 & 24.68 & 37.67 & 34.94 & 49.05 & 43.39 & 33.38 & 16.18 & 14.72 & 38.58 & 32.06 & 49.51 & 32.01 & 24.07 & 25.28 & 28.93 & 36.46 & 19.52 \\
Uniform & 42.48 & 51.98 & 47.80 & 57.92 & 60.45 & 47.63 & 23.59 & 24.38 & 54.57 & 48.72 & 59.52 & 44.10 & 35.14 & 8.21 & 43.24 & 52.00 & 20.95 \\
Isolated & 41.99 & 52.49 & 47.88 & 58.06 & 60.70 & 47.94 & 24.12 & 24.08 & 54.64 & 49.18 & 59.31 & 44.55 & 34.49 & 9.38 & 43.19 & 51.46 & 20.38 \\
Natural & 43.44 & 53.47 & 48.64 & 59.13 & 61.07 & 48.83 & 24.48 & 24.50 & 55.45 & 50.14 & 60.41 & 45.28 & 35.57 & 10.95 & 44.14 & 52.85 & 21.81 \\
MSL & 42.71 & 52.35 & 47.41 & 57.74 & 59.67 & 47.83 & 24.56 & 24.61 & 53.91 & 49.03 & 58.77 & 44.11 & 35.23 & 11.78 & 43.36 & 51.67 & 22.01 \\
\textsc{Climb} & 45.63 & 55.50 & 50.28 & 61.43 & 63.18 & 50.64 & 26.59 & 26.66 & 56.94 & 51.92 & 62.14 & 46.58 & 37.75 & 13.45 & 46.16 & 54.82 & 22.65 \\
\bottomrule
\end{tabular}}
\label{tab:flores_translation_detailed_results}
\end{table}

\begin{table}[htbp]
\centering
\caption{Detailed per-language performance on the \textbf{ARC-Challenge} benchmark (25-shot accuracy).}
\resizebox{\textwidth}{!}{%
\begin{tabular}{@{}lcccccccccccccccccc@{}}
\toprule
\textbf{Model / Method} & AR & DE & EN & ES & FR & ID & IT & JA & KO & MS & NL & PT & RU & TA & TH & TR & VI & ZH \\
\midrule
\multicolumn{19}{l}{\textbf{Open Source Multilingual LLMs}} \\
\midrule
LLaMA-3.2 & 26.64 & 30.93 & 42.26 & 35.16 & 33.42 & 30.34 & 32.88 & 28.67 & 31.21 & 30.76 & 30.30 & 32.88 & 29.53 & 25.12 & 28.69 & 29.05 & 30.45 & 32.80 \\
Qwen-3 & 34.00 & 42.03 & 54.39 & 43.51 & 41.31 & 41.22 & 43.49 & 35.42 & 36.33 & 37.58 & 38.32 & 43.11 & 39.92 & 28.09 & 32.77 & 33.03 & 37.52 & 45.54 \\
Gemma-3 & 25.58 & 29.77 & 38.41 & 30.69 & 31.03 & 30.27 & 30.27 & 27.92 & 28.42 & 28.18 & 27.09 & 30.94 & 28.42 & 25.92 & 25.92 & 27.59 & 27.26 & 29.94 \\
\midrule
\multicolumn{19}{l}{\textbf{Different Data Allocation Methods}} \\
\midrule
Uniform & 33.46 & 35.60 & 40.10 & 38.47 & 35.60 & 35.77 & 38.59 & 34.48 & 35.17 & 35.34 & 35.17 & 37.72 & \bf 38.23 & 22.78 & 32.00 & 35.85 & 35.09 & \bf 37.97 \\
Isolated & 31.19 & 35.53 & 38.72 & 37.02 & 36.25 & 37.45 & 37.87 & 35.44 & 34.62 & 37.02 & 36.00 & 37.26 & 34.75 & 23.15 & 31.71 & 34.62 & 32.72 & 34.71 \\
Natural & 31.75 & 33.98 & 38.37 & 35.91 & 34.60 & 34.76 & 36.61 & 33.24 & 32.80 & 33.63 & 33.50 & 35.84 & 33.81 & 21.99 & 30.05 & 33.69 & 33.48 & 35.72 \\
MSL & 32.31 & 34.59 & 38.90 & 36.64 & 35.30 & 35.46 & 37.34 & 33.74 & 33.20 & 34.12 & 33.98 & 36.38 & 34.32 & 22.60 & 30.70 & 34.47 & 34.28 & 36.74 \\
\textsc{Climb} & \bf 34.48 & \bf 37.10 & \bf 41.78 & \bf 39.05 & \bf 38.03 & \bf 38.27 & \bf 40.19 & \bf 36.33 & \bf 35.64 & \bf 37.20 & \bf 37.05 & \bf 39.67 & 37.40 & \bf 24.08 & \bf 32.54 & \bf 36.58 & \bf 36.40 & 36.30 \\
\bottomrule
\end{tabular}}
\label{tab:arc_challenge_detailed_results}
\end{table}

\begin{table}[htbp]
\centering
\caption{Detailed per-language performance on the \textbf{ARC-Easy} benchmark (25-shot accuracy).}
\resizebox{\textwidth}{!}{%
\begin{tabular}{@{}lcccccccccccccccccc@{}}
\toprule
\textbf{Model / Method} & AR & DE & EN & ES & FR & ID & IT & JA & KO & MS & NL & PT & RU & TA & TH & TR & VI & ZH \\
\midrule
\multicolumn{19}{l}{\textbf{Open Source Multilingual LLMs}} \\
\midrule
LLaMA-3.2 & 39.16 & 50.09 & 70.21 & 54.98 & 52.23 & 48.35 & 51.18 & 40.29 & 41.26 & 44.43 & 47.09 & 52.19 & 48.56 & 35.74 & 37.63 & 42.37 & 46.96 & 49.40 \\
Qwen-3 & 49.23 & 62.56 & 80.14 & 67.38 & 64.09 & 60.14 & 62.73 & 53.92 & 52.27 & 51.49 & 54.70 & 65.11 & 60.84 & 41.24 & 46.35 & 49.35 & 57.51 & 69.64 \\
Gemma-3 & 41.68 & 49.67 & 70.80 & 55.19 & 53.55 & 50.56 & 54.01 & 48.37 & 47.44 & 43.70 & 48.62 & 52.49 & 47.02 & 39.15 & 39.61 & 44.08 & 46.18 & 55.31 \\
\midrule
\multicolumn{19}{l}{\textbf{Different Data Allocation Methods}} \\
\midrule
Uniform & 56.70 & 62.68 & 70.93 & 67.27 & 63.81 & 64.28 & 64.07 & \bf 58.97 & \bf 58.01 & \bf 57.71 & \bf 62.59 & 66.34 & 60.24 & 29.64 & 49.59 & 60.20 & 58.76 & \bf 63.86 \\
Isolated & 55.12 & 62.07 & 69.93 & 65.59 & 63.45 & 63.74 & 61.73 & 56.77 & 56.98 & 56.93 & 61.73 & 63.21 & 58.62 & 30.29 & 48.31 & 59.46 & 56.56 & 63.08 \\
Natural & 53.88 & 59.60 & 67.83 & 63.25 & 61.68 & 61.16 & 60.30 & 53.99 & 53.45 & 52.40 & 57.84 & 62.68 & 57.09 & 29.54 & 47.44 & 56.40 & 55.10 & 60.12 \\
MSL & 55.21 & 60.93 & 68.99 & 64.61 & 63.06 & 62.34 & 61.63 & 55.14 & 54.64 & 53.74 & 59.17 & 64.14 & 58.49 & 30.41 & 48.57 & 57.73 & 56.42 & 61.55 \\
\textsc{Climb} & \bf 57.75 & \bf 63.84 & \bf 72.47 & \bf 67.74 & \bf 66.25 & \bf 65.45 & \bf 64.96 & 58.17 & 57.80 & 57.09 & 62.03 & \bf 67.11 & \bf 61.45 & \bf 32.24 & \bf 50.96 & \bf 60.64 & \bf 59.12 & 63.02 \\
\bottomrule
\end{tabular}}
\label{tab:arc_easy_detailed_results}
\end{table}


\begin{table}[htbp]
\centering
\caption{Detailed per-language performance on the \textbf{GPQA} benchmark (0-shot accuracy).}
\resizebox{\textwidth}{!}{%
\begin{tabular}{@{}lcccccccccccccccccc@{}}
\toprule
\textbf{Model / Method} & AR & DE & EN & ES & FR & ID & IT & JA & KO & MS & NL & PT & RU & TH & TL & TR & VI & ZH \\
\midrule
\multicolumn{19}{l}{\textbf{Open Source Multilingual LLMs}} \\
\midrule
LLaMA-3.2 & 23.54 & 25.67 & 26.39 & 23.30 & 26.39 & 23.30 & 22.81 & 22.81 & 25.20 & 23.79 & 23.30 & 23.30 & 24.02 & 23.79 & 25.67 & 23.30 & 23.79 & 24.50 \\
Qwen-3 & 27.47 & 31.60 & 32.53 & 28.06 & 32.79 & 31.89 & 31.60 & 30.38 & 29.79 & 27.72 & 34.87 & 31.60 & 32.53 & 28.95 & 31.89 & 31.30 & 28.36 & 31.60 \\
Gemma-3 & 24.58 & 23.30 & 25.45 & 23.03 & 24.44 & 23.49 & 25.25 & 23.49 & 22.76 & 24.91 & 23.49 & 24.91 & 22.05 & 24.58 & 21.25 & 22.76 & 26.60 & 25.45 \\
\midrule
\multicolumn{19}{l}{\textbf{Different Data Allocation Methods}} \\
\midrule
Uniform & 25.41 & 26.48 & 27.28 & 25.72 & 27.28 & 26.72 & 25.91 & 24.65 & \bf 26.24 & 24.90 & 27.52 & \bf 26.72 & 26.72 & 26.97 & 25.71 & 25.71 & 25.41 & 25.91 \\
Isolated & 23.51 & 22.77 & 26.27 & 21.76 & 24.24 & 25.03 & 25.03 & 23.51 & 24.51 & 25.24 & 25.24 & 25.03 & 23.27 & \bf 28.64 & 25.03 & 23.51 & 24.80 & 25.03 \\
Natural & 24.97 & 26.40 & 28.24 & 26.19 & 27.77 & 26.94 & 26.64 & 25.17 & 25.02 & 25.60 & 26.78 & 27.15 & 26.26 & 25.66 & 26.56 & 26.47 & 25.70 & \bf 27.16 \\
MSL & 24.23 & 25.43 & 27.06 & 25.34 & 26.61 & 25.94 & 25.64 & 24.19 & 24.00 & 24.72 & 25.89 & 26.27 & 25.10 & 24.44 & 25.50 & 25.39 & 24.53 & 26.20 \\
\textsc{Climb} & \bf 25.96 & \bf 27.09 & \bf 28.60 & \bf 27.24 & \bf 28.46 & \bf 27.71 & \bf 27.44 & \bf 25.88 & 25.71 & \bf 26.41 & \bf 27.59 & \bf 28.00 & 26.67 & 26.13 & \bf 27.28 & \bf 27.18 & \bf 26.20 & 26.98 \\
\bottomrule
\end{tabular}}
\label{tab:gpqa_detailed_results}
\end{table}


\begin{table}[htbp]
\centering
\caption{Detailed per-language performance on the \textbf{HellaSwag} benchmark (10-shot accuracy).}
\resizebox{\textwidth}{!}{%
\begin{tabular}{@{}lcccccccccccccccccc@{}}
\toprule
\textbf{Model / Method} & AR & DE & EN & ES & FR & ID & IT & JA & KO & MS & NL & PT & RU & TA & TH & TR & VI & ZH \\
\midrule
\multicolumn{19}{l}{\textbf{Open Source Multilingual LLMs}} \\
\midrule
LLaMA-3.2 & 36.12 & 42.69 & 67.10 & 47.45 & 46.86 & 43.14 & 45.06 & 36.78 & 37.02 & 40.66 & 43.35 & 46.28 & 42.38 & 35.30 & 35.21 & 36.46 & 42.35 & 43.48 \\
Qwen-3 & 39.37 & 45.66 & 65.36 & 51.17 & 50.93 & 46.08 & 48.86 & 42.25 & 39.52 & 42.49 & 43.30 & 51.26 & 45.53 & 35.81 & 36.79 & 36.01 & 44.79 & 52.39 \\
Gemma-3 & 35.91 & 40.54 & 58.37 & 42.84 & 45.18 & 42.15 & 43.81 & 37.61 & 36.51 & 38.84 & 41.17 & 44.27 & 39.03 & 34.94 & 33.76 & 34.87 & 38.20 & 41.35 \\
\midrule
\multicolumn{19}{l}{\textbf{Different Data Allocation Methods}} \\
\midrule
Uniform & 45.03 & 49.72 & 58.01 & 54.03 & 54.83 & 52.41 & 52.90 & 45.08 & 43.01 & 47.15 & 51.51 & 53.67 & 48.50 & 30.04 & 39.10 & 45.27 & 47.84 & \bf 48.06 \\
Isolated & 44.67 & 49.70 & 58.02 & 53.31 & 54.35 & 52.06 & 52.25 & 45.70 & 43.34 & 47.23 & 51.35 & 53.44 & 48.33 & 30.61 & 39.68 & 45.99 & 48.34 & 47.58 \\
Natural & 42.64 & 46.95 & 55.16 & 51.21 & 51.95 & 49.75 & 50.12 & 42.98 & 41.06 & 44.48 & 48.50 & 50.71 & 45.73 & 29.08 & 37.87 & 43.24 & 45.31 & 45.52 \\
MSL & 43.80 & 48.06 & 56.71 & 52.67 & 53.37 & 51.11 & 51.45 & 44.12 & 42.19 & 45.69 & 49.92 & 52.23 & 46.93 & 30.37 & 39.32 & 44.90 & 46.82 & 46.73 \\
\textsc{Climb} & \bf 45.71 & \bf 49.93 & \bf 58.99 & \bf 54.39 & \bf 55.31 & \bf 53.20 & \bf 53.38 & \bf 46.05 & \bf 44.05 & \bf 47.67 & \bf 51.79 & \bf 54.15 & \bf 48.65 & \bf 32.10 & \bf 40.95 & \bf 46.75 & \bf 48.51 & 45.92 \\
\bottomrule
\end{tabular}}
\label{tab:hellaswag_detailed_results}
\end{table}


\begin{table}[htbp]
\centering
\caption{Detailed per-language performance on the \textbf{TruthfulQA} benchmark (0-shot accuracy).}
\resizebox{\textwidth}{!}{%
\begin{tabular}{@{}lcccccccccccccccccc@{}}
\toprule
\textbf{Model / Method} & AR & DE & EN & ES & FR & ID & IT & JA & KO & MS & NL & PT & RU & TH & TL & TR & VI & ZH \\
\midrule
\multicolumn{19}{l}{\textbf{Open Source Multilingual LLMs}} \\
\midrule
LLaMA-3.2 & 38.66 & 37.41 & 34.59 & 37.20 & 36.56 & 38.92 & 34.16 & 36.91 & 39.59 & 36.09 & 37.20 & 36.76 & 40.04 & 36.76 & 36.32 & 37.90 & 42.73 & 41.36 \\
Qwen-3 & 49.16 & 50.27 & 47.94 & 50.01 & 50.54 & 48.24 & 50.01 & 51.56 & 47.18 & 47.70 & 46.15 & 52.98 & 51.17 & 47.35 & 42.46 & 46.15 & 52.03 & 48.96 \\
Gemma-3 & 39.20 & 41.60 & 39.60 & 39.83 & 38.73 & 42.47 & 40.92 & 37.39 & 41.14 & 37.83 & 36.28 & 42.03 & 44.24 & 40.05 & 34.51 & 39.38 & 43.80 & 42.25 \\
\midrule
\multicolumn{19}{l}{\textbf{Different Data Allocation Methods}} \\
\midrule
Uniform & 41.42 & 38.07 & 38.28 & 41.63 & 41.42 & 39.09 & 40.74 & 35.97 & \bf 41.42 & 39.09 & 39.73 & 39.94 & 39.94 & 38.07 & 37.02 & 38.28 & 41.42 & 41.63 \\
Isolated & 36.55 & \bf 42.71 & 37.18 & 39.93 & 40.55 & 39.49 & 40.13 & 37.59 & 40.78 & 36.98 & 39.73 & 41.43 & 38.48 & 38.26 & \bf 38.26 & 40.13 & \bf 45.00 & 41.61 \\
Natural & 41.62 & 40.17 & 40.95 & 42.12 & 41.83 & 40.91 & 40.76 & 38.60 & 40.09 & 39.40 & 40.42 & 41.94 & 40.99 & 39.06 & 37.51 & 39.74 & 42.35 & \bf 42.91 \\
MSL & 40.53 & 39.05 & 39.91 & 41.02 & 40.76 & 39.76 & 39.57 & 37.50 & 38.93 & 38.26 & 39.27 & 40.70 & 39.75 & 37.84 & 36.35 & 38.74 & 41.18 & 41.70 \\
\textsc{Climb} & \bf 42.05 & 40.57 & \bf 41.53 & \bf 42.57 & \bf 42.32 & \bf 41.36 & \bf 41.15 & \bf 38.99 & 40.49 & \bf 39.72 & \bf 40.80 & \bf 42.17 & \bf 41.09 & \bf 39.31 & 37.78 & \bf 40.38 & 42.62 & 42.03 \\
\bottomrule
\end{tabular}}
\label{tab:truthfulqa_detailed_results}
\end{table}


\section{Limitations and Future Work}
\label{appendix:limitations_future_work}

While our experiments demonstrate strong performance using the proposed multilingual allocation strategy based on scaling laws, several limitations should be acknowledged. First, our parametric fitting and allocation strategies are primarily validated on a 1.2 billion-parameter (1.2B) model, and although Section 4.2 indicates robust performance at a larger scale (7B), explicitly incorporating model size (\(N\)) into the allocation optimization could potentially yield even more optimal data distributions. Exploring how scaling laws evolve explicitly with both dataset size (\(D\)) and model scale (\(N\)) thus remains an open area for future research.

Secondly, our current methodology exclusively considers cross-lingual transfer between languages included within the training dataset. An important and intriguing direction for future work involves extending our approach to account for potential transfer effects to and from languages not directly represented in the training set. Such an extension would enable more comprehensive and strategically informed allocation decisions, optimizing not just for immediate languages but also for broader linguistic coverage and potential downstream adaptability.

\section{Social Impact}
\label{appendix:social_impact}

\textsc{Climb} contributes positively by systematically enhancing multilingual performance in large language models (LLMs), thereby significantly improving global accessibility to advanced AI capabilities across diverse linguistic communities. Such improvements have the potential to reduce linguistic biases, bridge language gaps, and enhance equitable information access globally. However, there remain potential risks, including inadvertent reinforcement of cultural or linguistic biases inherent in training data and the possibility of over-reliance on optimized multilingual models leading to reduced human oversight and critical evaluation. It is crucial to responsibly deploy \textsc{Climb}-optimized models with ongoing evaluation and monitoring, actively addressing ethical considerations and biases to ensure equitable and inclusive benefits.

\end{document}